\documentclass[preprint,11pt]{elsarticle}

\usepackage{graphicx}
\usepackage{multirow}
\usepackage{hhline}
\usepackage{glossaries}
\usepackage{booktabs}
\usepackage{tabularx}
\usepackage{array}
\usepackage{float}
\usepackage{setspace}

\usepackage[noend,ruled,linesnumbered]{algorithm2e}
\SetKw{Continue}{continue}

\newcolumntype{L}{>{\raggedleft\arraybackslash}X}

\usepackage[small, margin=1cm]{caption}
\usepackage{subcaption}

\usepackage{appendix}
\usepackage{color}
\definecolor{strcolor}{rgb}{0.6, 0.2, 0.6}
\definecolor{commentcolor}{rgb}{0.3125, 0.5, 0.3125}
\definecolor{keycol}{rgb}{0, 0, 1}

\usepackage{amssymb}
\usepackage{amsmath}
\usepackage{mathtools}
\usepackage{bm} 
\usepackage{bbm}

\usepackage{url}
\usepackage{ntheorem}

\theoremstyle{theorem}

\newtheorem{proposition}{Proposition}
\newtheorem{assumption}{Assumption}

\allowdisplaybreaks
\usepackage{natbib}
\bibpunct[, ]{(}{)}{,}{a}{}{,}%
\def\BIBand{and}%

\newcommand\given[1][]{\:#1\vert\:}
\newcommand{\norm}[1]{\left\lVert#1\right\rVert}

\newacronym{RO}{RO}{robust optimization}
\newacronym{CCP}{CCP}{chance-constrained programming}
\newacronym{SPR}{SPR}{stochastic programming with recourse}
\newacronym{VRP}{VRP}{vehicle routing problem}
\newacronym{CVRP}{CVRP}{capacitated vehicle routing problem}
\newacronym{CVRPSD}{CVRPSD}{capacitated vehicle routing problem with stochastic demand}
\newacronym{VRPTW}{VRPTW}{vehicle routing problem with time windows}
\newacronym{VRP-D}{VRP-D}{VRP with deadlines}
\newacronym{TSPTW}{TSPTW}{traveling salesman problem with time windows}
\newacronym{SAA}{SAA}{sample average approximation}
\newacronym{CSAA}{CSAA}{conditional sample average approximation}
\newacronym{RSAA}{RSAA}{residual-based sample average approximation}
\newacronym{FPF}{FPF}{feature projection function}
\newacronym{PTO}{PTO}{\emph{predict-then-optimize}}
\newacronym{LP}{LP}{linear program}
\newacronym{MIP}{MIP}{mixed integer program}
\newacronym{MILP}{MILP}{mixed integer linear program}
\newacronym{BPC}{BP\&C}{branch-price-and-cut}
\newacronym{CG}{CG}{column generation}
\newacronym{RMP}{RMP}{restricted master problem}
\newacronym{ESPPRC}{ESPPRC}{elementary shortest path problem with resource constraints}
\newacronym{RCSP}{RCSP}{resource-constrained shortest path}
\newacronym{DP}{DP}{dynamic programming}

\newcounter{relctr} 
\everydisplay\expandafter{\the\everydisplay\setcounter{relctr}{0}} 

\newcommand\labelrel[2]{%
  \begingroup
    \refstepcounter{relctr}%
    \stackrel{\textnormal{(\alph{relctr})}}{\mathstrut{#1}}%
    \originallabel{#2}%
  \endgroup
}
\AtBeginDocument{\let\originallabel\label} 

\begin{document}
	
	\linespread{1.6}\selectfont
	
	\begin{center}
		
		\vspace*{-0.5cm}
		
		\begin{huge}
			Contextual Stochastic Vehicle Routing\\
			with Time Windows
		\end{huge}
		
		\vspace*{0.6cm}
		
		{\setstretch{1.0}
			
			\textbf{Breno Serrano}$^*$\\
			{\footnotesize School of Management, Technical University of Munich, Germany, breno.serrano@tum.de}\\[10pt]
			
			\textbf{Alexandre M. Florio}\\
			{\footnotesize Amazon.com, Inc., Bellevue, United States, United States of America, aflorio@gmail.com}\\[10pt]
			
			\textbf{Stefan Minner}\\
			{\footnotesize School of Management and Munich Data Science Institute, Technical University of Munich, Germany, stefan.minner@tum.de}\\[10pt]
			
			\textbf{Maximilian Schiffer}\\
			{\footnotesize School of Management and Munich Data Science Institute, Technical University of Munich, Germany, schiffer@tum.de}\\[10pt]
			
			\textbf{Thibaut Vidal}\\[2pt]
			{\footnotesize D\'epartement de Math\'ematiques et de G\'enie Industriel, Polytechnique Montr\'eal, Canada, thibaut.vidal@polymtl.ca}\\
		}
	\end{center}
	
	\noindent
	\textbf{Abstract.} 
	We study the vehicle routing problem with time windows (VRPTW) and stochastic travel times, in which the decision-maker observes related contextual information, represented as feature variables, before making routing decisions. Despite the extensive literature on stochastic VRPs, the integration of feature variables has received limited attention in this context. We introduce the contextual stochastic VRPTW, which minimizes the total transportation cost and expected late arrival penalties conditioned on the observed features. Since the joint distribution of travel times and features is unknown, we present novel data-driven prescriptive models that use historical data to provide an approximate solution to the problem. We distinguish the prescriptive models between point-based approximation, sample average approximation, and penalty-based approximation, each taking a different perspective on dealing with stochastic travel times and features. We develop specialized branch-price-and-cut algorithms to solve these data-driven prescriptive models. In our computational experiments, we compare the out-of-sample cost performance of different methods on instances with up to one hundred customers. Our results show that, surprisingly, a feature-dependent sample average approximation outperforms existing and novel methods in most settings.
	
	\vspace*{0.2cm}
	
	\vspace*{-0.5cm}
	
	\noindent
	$^*$ Corresponding author
	
	\noindent
	Declarations of interest: none

	\linespread{2.0}\selectfont
	\setlength{\abovedisplayskip}{3pt}
	\setlength{\belowdisplayskip}{3pt}

\section{Introduction}

The \gls{VRP} is one of the most studied problems in operations research, with the majority of papers focusing on deterministic variants, in which the decision-maker has complete information regarding the model parameters. In practice, \glspl{VRP} have many sources of uncertainty, e.g., related to travel times, demands, or service times, which stimulated research interest in stochastic optimization models for \glspl{VRP}. However, many studies on stochastic \glspl{VRP} consider stylized problems and assume that the uncertain parameters of the optimization problem have known probability distributions.

With the growing availability of data, decision-makers can harness historical data on the uncertain parameters in addition to correlated contextual information, represented as feature variables, to improve uncertainty representation in \gls{VRP} models. The benefits of contextual optimization for decision-making problems under uncertainty are evident in works from many fields (see, e.g., \citealt{sadana2023survey}). However, within the \gls{VRP} literature, integrating feature variables related to uncertain parameters has received limited attention. We address this research gap and directly incorporate contextual information into the classical \gls{VRP} framework.

Specifically, we study the \gls{VRPTW} and uncertain travel times, in which the decision-maker observes related features before making routing decisions. For example, the decision-maker may use contextual information about road closures, seasonal events, or the day of the week to improve routing decisions. To capture the dependence of the optimization problem on external features, we introduce the \textit{contextual stochastic} \gls{VRPTW}, which minimizes the total transportation cost and expected late arrival penalties conditioned on a set of observed features. We present prescriptive models that use historical data to provide an approximate solution to the problem. We distinguish the data-driven prescriptive models between point-based approximation, \gls{SAA}, and penalty-based approximation, each taking a different perspective in dealing with uncertain travel times and features. In particular, point-based approximation models correspond to the surrogate problem in which travel times are fixed to some estimated values given, e.g., by using a predictive model. In contrast, \gls{SAA} models capture travel time variability by considering a set of travel time scenarios. In this setting, the decision-maker can take the observed features into account either by associating a feature-dependent weight function to each training observation or by directly constructing feature-dependent travel time scenarios. Finally, we present a penalty-based approximation model, which consists of a predictive model that is trained to directly predict late arrival penalties from features. We develop specialized \gls{BPC} algorithms and compare the prescriptive models in computational experiments.

\subsection{Related Work}

We identified two main research streams in the related literature. Specifically, we start with a review of related works on stochastic \glspl{VRP} and then discuss data-driven stochastic optimization problems that consider the presence of features for different problem settings beyond \glspl{VRP}.

\vspace*{0.5cm}

\noindent\textbf{Stochastic \glspl{VRP}.} Numerous works have investigated stochastic variants of the \gls{VRP}, e.g., with stochastic demands, travel times, or service times, among other sources of uncertainty. 
The three most common modeling approaches from the literature are based on \gls{RO}, \gls{CCP}, and two-stage \gls{SPR}. Similarly to our paper, many~\gls{SPR} formulations assume a simple recourse policy consisting of a penalty, e.g., for late or early arrivals when time windows are considered. For an introduction to stochastic \glspl{VRP}, we refer the reader to \cite{gendreau2014chapter} and \cite{OYOLA2017349,oyola2018stochastic}. 

The seminal work of \cite{laporte1992vehicle} introduced an uncapacited \gls{VRP} with stochastic service and travel times. \cite{kenyon2003stochastic} extended the work of~\cite{laporte1992vehicle} by focusing on minimizing the completion time instead of the total transportation cost. \cite{li2010vehicle} considered time window constraints, extending the formulations of \cite{laporte1992vehicle} to the stochastic \gls{VRPTW}, for which they proposed a heuristic based on tabu search. \cite{Ta__2014} later solved a \gls{VRPTW} under independently gamma-distributed travel times, in which early and late service are permitted but penalized. Their objective minimized the sum of transportation costs and the expected early and late arrival penalties. In contrast to~\cite{Ta__2014}, we make no assumptions about the distribution of travel times. In addition, we assume that early service is forbidden, which increases the problem difficulty because the distributions of service start times at customers are truncated (cf.~\citealt{zhang2019routing}).

From an \gls{RO} perspective, \cite{lee2012robust} and \cite{Adulyasak_2016} studied the stochastic \gls{VRP-D}, a special case of the \gls{VRPTW}. Further works by \cite{miranda2016vehicle} and \cite{errico2018vehicle} adopted a \gls{CCP} approach for the \gls{VRPTW} with stochastic travel and service times, respectively. \cite{Zhang_2021} addressed the \gls{VRPTW} with stochastic travel times and proposed a data-driven distributionally robust optimization model.

Focusing on the~\gls{CVRP} under demand uncertainty, \cite{gounaris2013robust} proposed an \gls{RO} formulation, and \cite{ghosal2020distributionally} and \cite{ghosal2021unifying} proposed distributionally robust chance-constrained optimization models. Recently, \cite{Rostami_2021} considered a \gls{CVRP} in which travel times are stochastic and statistically correlated. Their bi-objective model minimized the total expected travel time and total variance, assuming that the mean and covariance matrix of the travel times are known.

In terms of different optimization criteria, \cite{jaillet2016routing} proposed the \emph{requirements violation}~(RV) index, which considers both the probability and magnitude of time window violations. They model a \gls{TSPTW} where early service is permitted, and the total RV measure is minimized. Inspired by the RV index, \cite{zhang2019routing} proposed the \emph{essential riskiness} index, which enables a multi-commodity flow formulation of the \gls{TSPTW} where early service is forbidden.

A number of works implicitly considered the impact of external features by modeling time-dependent travel times. For a review on time-dependent routing problems, we refer the reader to \cite{gendreau2015time}. \cite{dabia2013branch} proposed an arc-based formulation for the \gls{VRPTW} under deterministic time-dependent travel times and \cite{tacs2014time} proposed metaheuristics for the \gls{VRPTW} with time-dependent and stochastic gamma-distributed travel times. In contrast to time-dependent \glspl{VRP}, we assume that travel times are time-independent within a service period, and we explicitly model the dependence of travel times on a set of external features.

Despite the extensive literature on stochastic \glspl{VRP}, previous works considered the presence of contextual information only implicitly, e.g., with time-dependent travel times. Incorporating contextual information into the model can significantly reduce routing costs. Accordingly, we contribute to the current state of the art by proposing novel models and algorithms for the contextual stochastic \gls{VRPTW}. In addition, we show how to adapt state-of-the-art \gls{BPC} techniques when contextual information is available.

\vspace*{0.5cm}

\noindent\textbf{Contextual stochastic optimization.} Many recent works have investigated the use of contextual information to improve decisions under uncertainty, with applications to transportation, inventory, and operations management, among others. The recent survey by~\cite{sadana2023survey} reviewed papers on contextual optimization with a focus on models and methods, while the review paper by~\cite{mivsic2020data} discussed recent applications in the operations management literature.

\citet{bertsimas2019predictive} proposed a framework for contextual optimization based on weighted \gls{SAA} and derived weight functions based on machine learning methods, such as $k$-nearest neighbors regression, local linear regression, and tree-based methods. The authors applied the framework to a two-stage shipment planning problem and to a multi-product inventory problem.
Nearly simultaneously, \cite{NEURIPS2018_f337d999} adapted the framework for problems with decision-dependent uncertainties. \citet{bertsimas2019predictions} and \citet{BERTSIMAS2023634} extended the framework of \citet{bertsimas2019predictive} to multi-stage and multi-period stochastic optimization problems, respectively. \cite{van2022robust} proposed robust formulations based on regularized \gls{SAA} for conditional two-stage optimization problems.
	
\citet{Elmachtoub_2021} proposed a framework tailored to optimization problems consisting of a linear objective with uncertain coefficients. They used feature data to learn a prediction model for the objective coefficients and adopted a modified loss function that directly leverages the structure of the optimization problem instead of minimizing a standard prediction error, such as the least squares loss. \citet{mandi2020smart} extended this framework to solve some discrete combinatorial problems and illustrated their algorithmic improvements on a weighted knapsack problem and a scheduling problem.

In the context of inventory optimization, \cite{beutel2012safety} and \cite{ban2019big} studied the data-driven newsvendor problem and proposed models for learning a decision function that predicts order quantities directly from feature observations. \cite{serrano2024bilevel} extended their framework, introducing an integrated approach based on bilevel programming that performs feature selection. \cite{mandl2023data} studied a multi-period commodity procurement problem under price uncertainty with forward and spot purchase options. They proposed a data-driven model to derive optimal purchase policies based on features, such as economic indicators.

In the context of last-mile delivery operations, \cite{Liu_2021} assigned orders to drivers under uncertain service times and considered that drivers' routes may deviate from planned routes. The authors used features, e.g., based on the customers' locations, to learn a prediction model for the drivers' travel times, which is then integrated into the order assignment optimization.

Motivated by the problem of forecasting electricity demands given a set of features, \cite{rios2015multi} proposed a scenario generation method based on estimating the distribution of forecast errors, i.e., residuals. \cite{ban2019dynamic} investigated a similar methodology for a multi-stage stochastic procurement problem under uncertain demand, where they learned a regression model that relates features to demands and used the residuals to generate demand samples for a new product. They subsequently solved an \gls{SAA} model to determine the optimal procurement policy.

Many works in the contextual optimization literature deal with optimization problems that can be efficiently solved, e.g., that can be modeled as linear programs. In contrast, the \gls{VRPTW} is a challenging optimization problem, for which many researchers have proposed problem-tailored solution techniques over the last decades. On the one hand, we show how existing contextual optimization methods can be applied to this challenging application. On the other hand, we investigate novel methods that can be generally applied to other contextual optimization problems.

\subsection{Contributions}
	
Our work contributes to the research streams outlined in the previous section. First, we introduce a novel formulation for the contextual stochastic \gls{VRPTW}, in which realizations of travel times are conditioned on a set of related features. Second, we show how existing contextual optimization methods can be applied to the contextual stochastic \gls{VRPTW} and propose novel data-driven methods leveraging historical data. In particular, we propose a \gls{CSAA} method, a general technique based on \gls{SAA} for solving contextual optimization problems. Third, we propose a problem-tailored penalty-based prescriptive method that uses machine learning to predict late arrival penalties at customers instead of relying on travel time estimates. Fourth, we describe a specialized \gls{BPC} algorithm that is generally applicable to stochastic \gls{VRP} variants in which the travel times can capture the problem uncertainty, e.g., when uncertain service times or costs can be modeled in terms of uncertain travel times. One key component of our \gls{BPC} implementation is the use of completion bounds instead of common dominance rules for discarding labels in the pricing algorithm. Fifth, we conduct computational experiments to compare the proposed prescriptive models and conclude that \gls{SAA} methods equipped with feature-dependent scenarios show the best average performance, providing solutions that are closest to the full-information benchmark.

The remainder of this paper is organized as follows. Section~\ref{sec:fbvrptwprobdef} formally introduces the contextual stochastic \gls{VRPTW}. Section~\ref{sec:dd-prescriptive-models} presents our data-driven prescriptive models. Section~\ref{sec:algoddvrptw} describes the corresponding solution methods based on \gls{BPC}. Section~\ref{sec:design-exp} presents our experimental design and Section~\ref{sec:results} discusses computational results. Finally, Section~\ref{sec:conclusion} concludes the paper.

\section{The Contextual Stochastic VRPTW}\label{sec:fbvrptwprobdef}

We define the contextual stochastic \gls{VRPTW} on a complete digraph $\mathcal{G}=(\mathcal{V}, \mathcal{A})$, where $\mathcal{V}=\{0,\ldots,N\}$ is the set of nodes and $\mathcal{A}=\{(i,j):i,j\in\mathcal{V}\}$ is the set of arcs. Set $\mathcal{V}\setminus\{0\}$ represents customers, and node $0$ denotes the depot. A homogeneous fleet of $K$ vehicles, each with a capacity of $Q$, is stationed at the depot. Each node $i\in\mathcal{V}$ has known demand $q_{i}$ and a time window $[e_{i},\ell_{i}]$, $0\leq e_{i}\leq\ell_{i}$. Each arc $(i,j)\in\mathcal{A}$ has a deterministic cost $c_{ij}$ and an uncertain travel time $\tilde{t}_{ij}$. Throughout the paper, we use a tilde to indicate random variables and follow standard boldfaced notation for vectors and vector-valued functions.

We focus on a setting in which the joint distribution of travel times $\mathbf{\tilde{t}}=[\tilde{t}_{ij}]_{(i,j)\in\mathcal{A}}$ is unknown. The decision-maker observes a set of features, represented by a $p$-dimensional vector $\mathbf{x}=[x_{1},\ldots,x_{p}]$, before the travel times are revealed. Features are exogenous variables that may be travel time predictors, e.g., day of the week, season, events, meteorological data, and road works or closures.

A route is a non-empty sequence of customers $\theta=(v_{1},\ldots,v_{L})$ such that $\sum_{i\in\theta}q_{i}\leq Q$ and $v_{i}\neq v_{j}$ if $i\neq j$. We further define:
\begin{equation}
	C_{\theta}=c_{0v_{1}}+\sum_{k=2}^{L}c_{v_{k-1}v_{k}}+c_{v_{L}0}
\end{equation}
as the transportation cost of route $\theta$. We assume that transportation costs are deterministic and independent of travel times and features, e.g., if drivers have fixed salaries, and if fuel consumption costs can be well approximated considering only the distance traveled. Note that this assumption is not restrictive since all proposed methods can be adapted to a setting with uncertain costs by using (simplified variants of) the methodology presented in this work.

All routes start at the depot. We assume that service times equal to 0, without loss of generality, as a positive service time at a customer can be added to the travel times of its outgoing arcs. Early service is forbidden, such that in case of early arrival at customer $i$, the vehicle must wait until $e_{i}$ before service can start. Given a route $\theta=(v_{1},\ldots,v_{L})$ and realized travel times $\mathbf{t}=[t_{ij}]_{(i,j)\in\mathcal{A}}$, we define the \textit{arrival time} $a_{\theta}(v_{k};\mathbf{t})$ at customer $v_{k}\in\theta$ as:
\begin{equation}
	a_{\theta}(v_{k};\mathbf{t})=\left\{
	\begin{array}{ll}
		t_{0v_{1}}, & \text{\ \ \ if }k=1,\\
		\max\{e_{v_{k-1}},a_{\theta}(v_{k-1};\mathbf{t})\}+t_{v_{k-1}v_{k}}, & \text{\ \ \ if }k\in\{2,\ldots,L\}. \\
	\end{array}
	\right.
\end{equation}

Since early arrivals are not permitted, the instant when customer $v_k\in\theta$ is served, i.e., the \emph{service start time} at customer $v_k$, is given by:
\begin{equation}
	s_{\theta}(v_k;\mathbf{t}) = \max\{e_{v_k},a_{\theta}(v_k;\mathbf{t})\}.
\end{equation}
Equivalently, $s_{\theta}(v_k; \mathbf{t}) + t_{v_k v_{k+1}} = a_{\theta}(v_{k+1}; \mathbf{t})$. Moreover, a penalty of $\pi(a_{\theta}(v_k;\mathbf{t})-\ell_{v_k})$ occurs in case of late arrivals at customer $v_k$, where $\pi(\cdot)$ is a nondecreasing penalty function such that $\pi(0)=0$. For notational convenience, we define $\pi(u)=0$ if $u<0$.

With $\Theta$ being the set of all routes, we represent a solution to the contextual stochastic \gls{VRPTW} by a vector ${\mathbf{z}=[z_{\theta}]_{\theta\in\Theta}\in\{0,1\}^{|\Theta|}}$, where $z_{\theta}=1$ if and only if route $\theta$ belongs to the solution. Accordingly, the feasible region is determined by the set:
\begin{equation}\label{eq:feasVTW}
	\mathcal{Z}_{\Theta} = \left\{\mathbf{z}\in\{0,1\}^{|\Theta|}\,\,\middle\vert\,\,\begin{aligned}
		&\sum_{\theta\in\Theta}\mathbbm{1}(i\in\theta)z_{\theta}=1,&i\in\mathcal{V}\setminus\{0\},&\qquad\text{(a)}\\
		&\sum_{\theta\in\Theta}z_{\theta}\leq K&&\qquad\text{(b)}
	\end{aligned}\right\},
\end{equation}
where $\mathbbm{1}(\cdot)$ is the indicator function. The partitioning constraints (\ref{eq:feasVTW}a) ensure that all customers are visited once, and constraint (\ref{eq:feasVTW}b) enforces that no more than $K$ vehicles are used.
	
We formulate the contextual stochastic \gls{VRPTW} as a two-stage stochastic program: 
\begin{gather}
	\mathbf{z}^*(\mathbf{x}) = \arg \min_{\mathbf{z}\in\mathcal{Z}_{\Theta}} \,\, C(\mathbf{z}) + \mathbb{E} \left[Q(\mathbf{z}, \mathbf{\tilde{t}}) \,\middle|\, \mathbf{\tilde{x}}=\mathbf{x} \right], \label{model:vtw} \tag{CS-VRPTW}
\end{gather}
where the first-stage objective corresponds to the total transportation costs $C(\mathbf{z}) = \sum_{\theta\in\Theta} C_\theta z_\theta$, and the second-stage recourse policy consists of penalizing late arrivals, leading to the second-stage value function:
\begin{equation}
	Q(\mathbf{z}, \mathbf{t}) = \sum_{\theta\in\Theta}  \Big(\sum_{i\in\theta} \pi(a_{\theta}(i; \mathbf{t}) -\ell_i)\Big) z_{\theta}. \label{eq:sec-stage-value}
\end{equation}
We further denote by $f_{\Theta}(\mathbf{z,t})$ the value of a solution~$\mathbf{z}$ under realized travel times~$\mathbf{t}$:
\begin{equation}
	f_{\Theta}(\mathbf{z},\mathbf{t}) = C(\mathbf{z}) + Q(\mathbf{z,t}). 
\end{equation}
Thus, we can equivalently express the~\ref{model:vtw} as:
\begin{equation}
	\mathbf{z}^*(\mathbf{x}) = \arg \min_{\mathbf{z}\in\mathcal{Z}_{\Theta}} \,\, \mathbb{E} \left[f_{\Theta}(\mathbf{z}, \mathbf{\tilde{t}}) \,\middle|\, \mathbf{\tilde{x}}=\mathbf{x} \right]. 
\end{equation}
	
Since the joint distribution of travel times and features is unknown, we approximate the conditional expectation in the~\ref{model:vtw} formulation using data-driven prescriptive models that leverage historical travel times and features.


\section{Data-driven Prescriptive Models}\label{sec:dd-prescriptive-models}

As the contextual stochastic \gls{VRPTW} formulation cannot be solved directly, we discuss approximate and computationally tractable reformulations in the following. 
We assume that the decision-maker has access to historical travel times for the last $n$ periods, represented by the $n{\times}|\mathcal{A}|$ matrix ${\mathbf{T}=[\mathbf{t}^{1},\ldots,\mathbf{t}^{n}]^{\top}}$, where $\mathbf{t}^{k}=[t_{ij}^{k}]_{(i,j)\in\mathcal{A}}$ are the travel times observed at period~$k$. Further, the decision-maker has access to feature data for the last $n$ periods, represented by the $n{\times}p$ matrix $\mathbf{X}=[\mathbf{x}^{1},\ldots,\mathbf{x}^{n}]^{\top}$, where $\mathbf{x}^{k}=[x_{1}^{k},\ldots,x_{p}^{k}]$ contains the realizations of $p$ features at period~$k$. We assume w.l.o.g. that $x_{1}^{k}=1$ for all $k\in\{1,\ldots,n\}$. We further assume that the data is low-dimensional, i.e., $p<n$, and that $\mathbf{X}$ has full rank. 
The decision-maker solves the \ref{model:vtw} for period $n+1$, having observed the feature vector $\mathbf{x}^{n+1}$ but without knowing the realization of travel times.


\subsection{Point-based Approximation}\label{sec:vtwdet}
We simplify the~\ref{model:vtw} by fixing the travel times to point estimates ${\mathbf{\hat{t}}=[\hat{t}_{ij}]_{(i,j)\in\mathcal{A}}}$ and considering the deterministic optimization problem:
\begin{align*}
	\mathbf{\hat{z}}^*_{\textsc{d}}(\mathbf{x}) \, &= \, \arg \min_{\mathbf{z}\in\mathcal{Z}_{\Theta}} \,\, f_{\Theta} \left(\mathbf{z},\mathbf{\hat{t}}\right) \label{model:pvtw}\tag{P} \, = \, \arg \min_{\mathbf{z}\in\mathcal{Z}_{\Theta}} \,\, C(\mathbf{z}) + Q(\mathbf{z,\hat{t}}).
\end{align*}

In this setting, different options to compute $\mathbf{\hat{t}}$ from historical travel times $\mathbf{T}$ and feature data $\mathbf{X}$ exist. If $\mathbf{X}$ is a poor predictor of $\mathbf{T}$, then a reasonable option is to set $\hat{t}_{ij}$ to a statistic of $\{t_{ij}^{k}\}_{k\in\{1,\ldots,n\}}$, e.g., the mean, median, or a higher percentile in case more protection against late arrivals is desired. Otherwise, if $\mathbf{X}$ is a good predictor of $\mathbf{T}$, we may fix travel times to:
\begin{equation}\label{eq:vrptwpto}
	\mathbf{\hat{t}}=\mathbf{g}(\mathbf{x}^{n+1};\bm{\varphi}),
\end{equation}
where $\mathbf{g}:\mathbb{R}^{p}\mapsto\mathbb{R}^{|\mathcal{A}|}$ is a travel time prediction model, e.g., a multivariate regression model or a neural network, with parameters $\bm{\varphi}$ trained from historical travel times and features. The latter approach corresponds to the \gls{PTO} paradigm \citep{Elmachtoub_2021}. Since point estimates are used, Model~\eqref{model:pvtw} does not capture the impact of travel time variability on the decision cost and thus may yield poor decisions because the effects of under- and over-predicting travel times are not symmetric.

Note that a special case of Model~\eqref{model:pvtw} arises when the penalty function is such that $\pi(s)=\infty$ if $s>0$. In this case, the model reduces to the classical \gls{VRPTW} with hard time windows, for which efficient exact algorithms exist (cf. \citealt{pessoa2020generic}). For general non-decreasing penalty functions, we solve Model~\eqref{model:pvtw} with a tailored \gls{BPC} algorithm as described in Section \ref{sec:algoddvrptw}.


\subsection{Sample Average Approximation}\label{sec:vtwstoch}
Travel time variability plays a key role in the \ref{model:vtw} since late arrival penalties grow with the amount of delay.
We represent travel time variability in the stochastic model with a set of scenarios $\{\mathbf{t}^{\omega}\}_{\omega\in\Omega}$, where $\mathbf{t}^{\omega}=[t_{ij}^{\omega}]_{(i,j)\in\mathcal{A}}$ is a possible realization of travel times at period~${n+1}$, corresponding to scenario $\omega\in\Omega$. We further associate a feature-dependent weight function ${\alpha^\omega(\mathbf{x}): \mathbb{R}^p \mapsto \mathbb{R}}$ to each scenario $\omega\in\Omega$ and approximate the conditional expected penalty using weighted \gls{SAA}:
\begin{align*}
	\mathbf{\hat{z}}^*_{\textsc{s}}(\mathbf{x}) \, &= \, \arg \min_{\mathbf{z}\in\mathcal{Z}_{\Theta}} \,\, \sum_{\omega\in\Omega} \alpha^{\omega} f_{\Theta} \left(\mathbf{z},\mathbf{t}^{\omega}\right)\label{model:svtw}\tag{S} \, = \, \arg \min_{\mathbf{z}\in\mathcal{Z}_{\Theta}} \,\, C(\mathbf{z}) + \sum_{\omega\in\Omega} \alpha^{\omega} Q(\mathbf{z},\mathbf{t}^\omega),
\end{align*}
where we assume, w.l.o.g., that $\sum_{\omega\in\Omega}\alpha^\omega = 1$.
Model~\eqref{model:svtw} computes the second-stage cost by \gls{SAA} \citep{shapiro2014lectures}. Note that Model~\eqref{model:pvtw} is a special case of Model~\eqref{model:svtw} with $|\Omega|=1$. 

Before solving Model~\eqref{model:svtw}, the decision-maker must define the set of scenarios~$\Omega$. If the feature data $\mathbf{X}$ are poor predictors of the travel times~$\mathbf{T}$, then a sensible option is to define the travel time scenarios as the historical travel times $\{\mathbf{t}^\omega\}_{\omega\in\Omega} = \{\mathbf{t}^k\}_{k\in\{1,\dots,n\}}$, with $\alpha^\omega=1/n$ for $\omega\in\Omega$. 


\subsection{Conditional SAA}\label{sec:fbsaa}

To generate feature-dependent scenarios, we specify a probabilistic model for travel times and features. Let $\mathbf{\tilde{x}}=[\tilde{x}_{1},\ldots,\tilde{x}_{p}]$ be a random vector representing features before they are revealed at the current period $n+1$. Then, we assume that travel times (conditioned on~$\mathbf{\tilde{x}}$) are given by:
\begin{equation}\label{eq:ttmodel}
	\mathbf{\tilde{t}}\given\mathbf{\tilde{x}}=\mathbf{B}^{\top}\mathbf{\tilde{x}}+\bm{\tilde{\varepsilon}},
\end{equation}
where $\mathbf{B}$ is a $p{\times}|\mathcal{A}|$ matrix of constant parameters and $\bm{\tilde{\varepsilon}}=[\tilde{\varepsilon}_{ij}]_{(i,j)\in\mathcal{A}}$ is a random noise vector following a multivariate distribution with zero mean and unknown covariance matrix~$\bm{\Sigma}$.

The conditional travel times model from Equation \eqref{eq:ttmodel} generalizes the correlated travel times model by \cite{jaillet2016routing} in that $\tilde{x}_{1},\ldots,\tilde{x}_{p}$ are not assumed to be independent and the noise terms $\tilde{\varepsilon}_{ij}$ allow for a stronger or weaker dependence between $\tilde{t}_{ij}$ and $\mathbf{\tilde{x}}$ for each arc. Therefore, the feature space may include interaction terms between travel time covariates, which may improve travel time prediction accuracy as non-linearities can be captured. Moreover, the error terms $\tilde{\varepsilon}_{ij}$ may be correlated, so the model allows for correlated travel time variability caused by unanticipated events, e.g., emergency work or traffic incidents.

From historical travel times $\mathbf{T}$ and feature data $\mathbf{X}$, we estimate matrix $\mathbf{B}$ by least squares:
\begin{equation}
	\mathbf{\hat{B}}=(\mathbf{X}^{\top}\mathbf{X})^{-1}\mathbf{X}^{\top}\mathbf{T}.
\end{equation}
Following multivariate analysis theory \citep{Rencher_2012}, an estimate of the covariance matrix $\bm{\Sigma}$ is given by:
\begin{equation}
	\bm{\hat{\Sigma}}=\frac{\mathbf{T}^{\top}\mathbf{T}-\mathbf{\hat{B}}^{\top}\mathbf{X}^{\top}\mathbf{T}}{n-p}.
\end{equation}

Accordingly, if the feature data are good travel time predictors, we obtain a feature-dependent representation of travel time uncertainty at period~${n+1}$ by sampling travel times from a multivariate distribution with mean and covariance given by $\mathbf{\hat{B}}^{\top}\mathbf{x}^{n+1}$ and $\bm{\hat{\Sigma}}$, respectively.

\subsection{Residual-based SAA}\label{sec:rsaa}

The conditional \gls{SAA} model from Section~\ref{sec:fbsaa} requires the decision-maker to make distributional assumptions on travel times to generate feature-dependent scenarios. We present a distribution-free \gls{RSAA} model based on the residual tree method of \cite{ban2019dynamic}. 
Consider a travel time prediction model $\mathbf{g}(\cdot)$ as introduced in Equation~\eqref{eq:vrptwpto} with parameters learned from historical travel times $\mathbf{T}$ and features $\mathbf{X}$.
We compute the prediction error, which we denote as the \textit{residual}, associated with training observation $k$ as:
\begin{equation}
	\boldsymbol{\epsilon}^k = \mathbf{g}(\mathbf{x}^k) - \mathbf{t}^k.
\end{equation}

Given a feature vector $\mathbf{x}^{n+1}$ at period $n+1$, we generate a set of feature-dependent scenarios where scenario $k$ is given by the sum of the model prediction for the observed feature vector and the residual associated with training observation $k$:
\begin{equation}
	\Omega = \left\{\mathbf{g}(\mathbf{x}^{n+1}; \bm{\varphi}) + \boldsymbol{\epsilon}^k \right\}_{k=1,\dots, n}.
\end{equation}
We then solve Model~\eqref{model:svtw} with uniformly weighted residual-based scenarios $\Omega$.

\subsection{Late Arrival Penalty Approximation}\label{sec:vtwpresc}

Instead of leveraging feature data to estimate the travel times distribution, and then using such estimates to approximate the expected penalty of a solution, we can use data to predict penalties directly. We define the penalty-based approximation model: 
\begin{align*}
	\mathbf{\hat{z}}^*_{\textsc{p}}(\mathbf{x}) \, &= \, \arg \min_{\mathbf{z}\in\mathcal{Z}_{\Theta}} \hat{f}_{\Theta} (\mathbf{z}, \mathbf{x}) \label{model:lvtw}\tag{L} \,= \, \arg \min_{\mathbf{z}\in\mathcal{Z}_{\Theta}} C(\mathbf{z}) + \hat{Q}(\mathbf{z,x}), 
\end{align*}
where $\hat{Q}(\mathbf{z,x})$ approximates the conditional expected penalty $\mathbb{E} \left[ Q(\mathbf{z}, \mathbf{\tilde{t}}) \,\middle|\, \mathbf{\tilde{x}} = \mathbf{x}\right]$.

By definition, the conditional expected penalty of a solution $\mathbf{z}$ in Equation~\eqref{eq:sec-stage-value} can be decomposed into the sum of the late arrival penalties at the customers along the routes in $\mathbf{z}$. Clearly, the predicted penalty at a given customer~$i$ along a given route~$\theta\in\Theta$ depends on customer and route characteristics, e.g., the expected penalty is correlated with the position of $i$ in the route or the customer time window, in addition to the features $\mathbf{x}$ related to the travel times. Therefore, we define a \mbox{\gls{FPF}}, ${\mathbf{f}:\mathbb{R}^{p}\times\Theta\times\mathcal{V}\setminus\{0\}\mapsto\mathbb{R}^{\bar{p}}}$, that combines travel time covariates with node and route characteristics to obtain a vector of penalty predictors $\mathbf{y}\in\mathbb{R}^{\bar{p}}$:
\begin{equation}
	\mathbf{y} = \mathbf{f}(\mathbf{x}, \theta, i).
\end{equation}
Essentially, $\mathbf{f}(\mathbf{x},\theta,i)$ projects $\mathbf{x}$ onto a $\bar{p}$-dimensional feature space with potential predictors of the penalty at customer $i\in\theta$.
We describe the projected features in detail in~\ref{sec:fpf}.

A prediction model $h(\cdot)$ outputs late arrival penalty predictions based on the projected features~$\mathbf{y}$ given by the \gls{FPF}. We approximate the expected value $\hat{Q}$ of the second-stage penalty as:
\begin{equation}
	\hat{Q}(\mathbf{z}, \mathbf{x}) = \sum_{\theta\in\Theta} \sum_{i \in \theta} h(\mathbf{y};\mathbf{H}) \, z_{\theta}
\end{equation}
where $\mathbf{H}$ denotes the trained parameters of the prediction model $h(\cdot)$.

We adopt a supervised learning setting and build a training dataset with $\bar{n}$ observations of the projected features $\mathbf{y}^k$ and penalties $\pi^k$, for $k\in\{1,\dots,\bar{n}\}$. Specifically, we generate the projected features, represented by $\mathbf{Y} = [\mathbf{y}^1,\dots, \mathbf{y}^{\bar{n}}]$, from the original feature data~$\mathbf{X}$, historical travel times~$\mathbf{T}$, and a set of routes $\Theta$. In our experiments, we construct a set~$\Theta$ of routes by solving the other prescriptive models, i.e., point-based and \gls{SAA} models, using the \gls{BPC} algorithm described in Section~\ref{sec:algoddvrptw}, and record all routes evaluated during the solution process. The corresponding penalties, represented by $\boldsymbol{\pi}=[\pi^1,\dots,\pi^{\bar{n}}]$, can be computed by propagating the arrival times at each node along the route based on the respective travel times. The learning problem then corresponds to finding the model parameters that minimize some empirical risk of the form:
\begin{equation}
	\mathbf{\hat{H}} = \arg \min_{\mathbf{H}} \, \sum_{k=1}^{\bar{n}} \mathcal{L}(h(\mathbf{y}^k; \mathbf{H}), \pi^k)
\end{equation}
where $\mathcal{L}$ defines a loss function, e.g., a squared error, and the prediction model can be any supervised regression model, e.g., based on linear regression or neural networks.


\section{Solution Methods for the Data-driven Prescriptive Models}\label{sec:algoddvrptw}

We adopt a set-partitioning formulation, where the set of decision variables representing the feasible routes~$\Theta$ can be exponentially large. To solve the proposed models, we rely on branch-and-price, which is an effective method for solving integer linear programs with a very large number of variables \citep{Barnhart_1998}. In vehicle routing, the technique is applied to solve many problem variants that admit a set-partitioning formulation \citep{Costa_2019}. We start by introducing the solution method for solving Model~\eqref{model:svtw} in Sections~\ref{sec:column-generation} and \ref{sec:pricing} before we explain how to adapt the method to solve Model~\eqref{model:lvtw} in Section~\ref{sec:algo-penalty-based}.

\subsection{Column Generation}\label{sec:column-generation}
In branch-and-price, we solve the continuous relaxation of the set-partitioning formulation by column generation. We define a \gls{RMP} considering only a subset $\Theta'\subset\Theta$ of the set of feasible routes, which yields the following linear program:
\begin{align*}
	\min   \quad & \sum_{\omega\in\Omega}\alpha^\omega f_{\Theta'}\left(\mathbf{z},\mathbf{t}^\omega\right) &\label{model:rmp}\tag{RMP}\\
	\text{s.t.}   \quad & \sum_{\theta\in\Theta'}\mathbbm{1}(i\in\theta)z_{\theta}=1, &\qquad i\in\mathcal{V}\setminus\{0\} \\
	&\sum_{\theta\in\Theta'}z_{\theta}\leq K\\
	& 0 \leq z_{\theta} \leq 1, & \qquad \theta \in \Theta'.
\end{align*}

The column generation procedure repeatedly solves \eqref{model:rmp}, identifies routes that correspond to negative reduced cost variables, and adds those routes to $\Theta'$. To this end, we solve the corresponding pricing problem: 
\begin{gather}
\min_{\theta\in\Theta} \,\,\,\, \overline{C}_{\theta}\triangleq C_{\theta}+\sum_{\omega\in\Omega}\alpha^\omega\sum_{i\in\theta}\pi(a_{\theta}(i;\mathbf{t}^\omega)-\ell_{i})-\sum_{i\in\theta}\gamma_{i}-\mu,\label{model:pp-svtw}\tag{PP-S}
\end{gather}
where, given a solution to \eqref{model:rmp}, $\bm{\gamma}=[\gamma_{i}]_{i\in\mathcal{V}\setminus\{0\}}$ and $\mu$ are the dual values associated with Constraints~(\ref{eq:feasVTW}a) and (\ref{eq:feasVTW}b), respectively. The pricing problem is an \gls{ESPPRC}, which is $\mathcal{NP}$-hard.

\subsection{Pricing Algorithm}\label{sec:pricing}
We use an extend-and-bound labeling algorithm, where a label represents a partial path from the depot to a customer. The labeling procedure creates new labels by extending partial paths to all feasible customers. For each new label extension, we compute a lower bound on the reduced cost of all routes that can be generated from the label extension, i.e., a completion bound. Then, we discard the label extension if this bound is non-negative.

A label $L_{\theta}$ representing a path $\theta$ is a tuple $L_{\theta}=(i,\overline{C}_{\theta},q_{\theta},\tau_\theta)$, where $i\in\mathcal{V}\setminus\{0\}$ is the last customer in~$\theta$, $\overline{C}_{\theta}$ is the reduced cost, $q_{\theta}=\sum_{v\in\theta}q_v$ is the cumulated load along $\theta$, and $\tau_\theta$ is the earliest time at which service can start at customer $i$ if $\theta$ is used to reach $i$: 
\begin{equation}
	\tau_\theta = \min_{\omega\in\Omega}\left\{ s_{\theta}(i; \mathbf{t}^\omega) \right\}, \label{eq:label-tau}
\end{equation}
considering all scenarios $\omega\in\Omega$.

If we extend label $L_{\theta}$ along an arc $(i,j)$, we generate a new path $\theta' = \theta\oplus (i,j)$ with a corresponding label $L_{\theta'}=(j, \overline{C}_{\theta'}, q_{\theta'}, \tau_{\theta'})$, such that:
\begin{align}
	\overline{C}_{\theta'} &= \overline{C}_{\theta} - c_{i0} + c_{ij} + c_{j0} + \sum_{\omega\in\Omega} \alpha^{\omega} \cdot \pi(a_{\theta'}(j; \mathbf{t}^{\omega}) - \ell_j ) - \gamma_j \label{eq:label-ext-cost} \\
	\tau_{\theta'} &= \min_{\omega\in\Omega}\{s_{\theta'}(j;\mathbf{t}^\omega) \} \label{eq:label-ext-tau}.
\end{align}
Path $\theta'$ is feasible if $q_{\theta'}\leq Q$. Otherwise, it is infeasible, and we discard the new label.

A completion bound $\widehat{T}(i,q)$ is a lower bound on the value of the best path starting at customer~$i$ and ending at the depot with a total load less than or equal to $q$. If the following condition holds, then we can discard label $L_\theta$ without losing any negative reduced cost route:
\begin{equation}
	\overline{C}_{\theta}+\widehat{T}(i,Q-q_{\theta})\geq 0.
\end{equation}
Since we solve the pricing problem multiple times during the execution of the \gls{BPC} method, it is important to generate completion bounds quickly. Hence, we employ heuristic algorithms based on a \gls{RCSP} problem and based on a knapsack formulation (cf. \citealt{florio2020new}). Unlike many labeling algorithms from the literature, we do not discard labels based on dominance rules. In order to use dominance rules in our setting, we would need labels to store a resource corresponding to the service start time under each travel time scenario. We leave the investigation of dominance rules and the assessment of their effectiveness for future research.

\vspace*{0.5cm}

\noindent\textbf{Resource-Constrained Shortest Path (RCSP) Bound.} 
We derive \gls{RCSP} bounds by relaxing the elementary requirement of the \gls{ESPPRC} and enforcing 2-cycle elimination, i.e., we allow paths with cycles only if cycles have a length of at least three customers. Resource constraints correspond to the capacity constraints on the vehicles. We compute a lower bound on the late arrival penalties along a path~$\theta$ based on the earliest service start time $\tau_\theta$. We provide a \gls{DP} algorithm for computing the \gls{RCSP} bounds in~\ref{appendix-RCSP-bound}, where we apply a time discretization with a time step $\Delta t$ such that we can efficiently store the optimal value of each subproblem. The \gls{DP} algorithm solves the following recursive formulation:
\begin{equation}
	\widehat{T}_{\textsc{RCSP}}(i,Q-q_\theta) = - c_{i0} + \min_{\substack{j\in\mathcal{V}\setminus\{0\} : \,\, \theta'=\theta \oplus (i,j), \\
	j\neq i, \, \rho(i) \neq j, \\ q_\theta+q_j\leq Q }} \left\{ c_{ij} + \pi(\delta_{\theta'} - \ell_j) - \gamma_j + c_{j0} + \widehat{T}_{\textsc{RCSP}}(j,Q-q_{\theta'}) \right\}, \label{eq:rcsp-bound-def}
\end{equation}
where $\theta' = \theta\oplus (i,j)$ denotes the extension of path $\theta$ along the arc $(i,j)$ and $\delta_\theta$ is a multiple of $\Delta t$ representing a lower bound on the service start time of customer $i\in\theta$, such that $\tau_{\theta} \in [\delta_\theta, \delta_\theta+\Delta t]$. The resulting completion bound expresses the maximum reduced cost decrease from node $i$ given that departure from node $i$ is not before $\delta_\theta$.

\begin{proposition}\label{prop:completion-bound-rcsp}
	Let $\theta$ be a route starting at the depot and ending at customer $i$ with reduced cost $\overline{C}_{\theta}$ and cumulative demand $q_{\theta}$. 
	Let $L_{\theta'}$ be a label extension associated with the route $\theta'=\theta \oplus\mathcal{E}$, where $\mathcal{E}=(u_1, u_2, \dots, u_L)$ is a path such that $u_j \notin\theta,\, \forall u_j \in\mathcal{E}$, and $q_{\theta'}\leq Q$. 
	If $\overline{C}_\theta + \widehat{T}_{\textsc{RCSP}}(i, Q-q_\theta) \geq 0$, then the label extension has non-negative reduced cost, i.e., $\overline{C}_{\theta'} \geq 0$.
\end{proposition}
\textit{Proof.}\hspace*{0.2cm} We provide proof for the \gls{RCSP} bound in~\ref{sec:proof-completion-bounds}. 
\hfill$\square$

\vspace*{0.5cm}

\noindent\textbf{Knapsack Bound.} 
Consider a label~$L_{\theta}$ representing a route $\theta$ that ends at customer $i$. We build a \mbox{$\{0,1\}$-knapsack} problem with $N$ items and capacity $Q-q_{\theta}$. We associate a weight $w_j=q_j$ to each item $j\in\{1,\dots,N\}$, equal to the demand of customer $j\in\mathcal{V}\setminus\{0\}$, and a value consisting of the dual value $\gamma_j$ minus a lower bound on the average penalty:
\begin{equation}
	v_{ij}(\theta) = \gamma_j - \pi(\tau_\theta+\min_{\omega\in\Omega}t^\omega_{ij}-\ell_j),
\end{equation}
such that the following inequality holds: 
\begin{equation}
	- v_{ij}(\theta) \,\leq\, \sum_{\omega\in\Omega} \alpha^{\omega} \cdot \pi(a_{\theta\oplus (i,j)}(j; \mathbf{t}^{\omega}) - \ell_j ) - \gamma_j. \label{knapsack-value-bound}
\end{equation}
We can express the knapsack problem as an integer linear program:
\begin{align}
	\max \quad &\sum_{j\in\mathcal{V}\setminus\{0\}} v_{ij}(\theta) \, z_j \label{start-knapsack}\\ 
	\text{s.t.}\quad & \sum_{j\in\mathcal{V}\setminus\{0\}} q_j z_j \leq Q-q_{\theta}\\
	& z_j =0 \text{ if } j\in\theta & \forall j \in \mathcal{V}\setminus\{0\} \label{knapsack-constr2}\\
	& z_j \in \{0,1\} & \forall j \in \mathcal{V}\setminus\{0\} \label{end-knapsack}
\end{align}
where Constraints~\eqref{knapsack-constr2} enforce that item $j$ is considered for inclusion in the knapsack only if it is not already in route $\theta$. 
Let~$\mathbf{z}^*$ be an optimal solution to the $\{0,1\}$-knapsack problem and let $\overline{\mathbf{z}}$ be a solution to its linear relaxation. 
Then, we define the following completion bound:
\begin{equation}
	\widehat{T}_{ks}(i,Q-q_\theta) = -\sum_{j\in\mathcal{V}\setminus\{0\}} v_{ij}(\theta) \, \overline{z}_{j}.
\end{equation}

\begin{proposition}\label{prop:completion-bound-knapsack}
	Let $\theta$ be a route starting at the depot and ending at customer $i$ with reduced cost $\overline{C}_{\theta}$ and cumulative demand $q_{\theta}$. 
	Let $L_{\theta'}$ be a label extension associated with the route $\theta'=\theta \oplus\mathcal{E}$, where $\mathcal{E}=(u_1, u_2, \dots, u_L)$ is a path such that $u_j \notin\theta,\, \forall u_j \in\mathcal{E}$, and $q_{\theta'}\leq Q$. 
	If $\overline{C}_\theta + \widehat{T}_{ks}(i, Q-q_\theta) \geq 0$, then the label extension has non-negative reduced cost, i.e., $\overline{C}_{\theta'} \geq 0$.
\end{proposition}
\textit{Proof.}\hspace*{0.2cm} We provide proof for the knapsack bound in~\ref{sec:proof-completion-bounds}. 
\hfill$\square$

\subsection{Column Generation for Penalty-based Approximation}\label{sec:algo-penalty-based}

We now describe how the \gls{CG} approach from the previous sections can be modified to solve the penalty-based approximation of Model~\eqref{model:lvtw}. The \gls{RMP} for the penalty-based model optimizes the objective function:
\begin{equation}
	\min \,\,\, \hat{f}_{\Theta'}(\mathbf{z}, \mathbf{x}),\label{model:rmp-lvtw}\tag{RMP-L}
\end{equation}
while the constraints remain the same as in~\eqref{model:rmp}. Accordingly, we incorporate the late arrival penalty prediction model into the pricing problem:
\begin{gather}
	\min_{\theta\in\Theta} \,\,\,\, \overline{C}_{\theta}\triangleq C_{\theta}+\sum_{i\in\theta}h(\mathbf{f}(\mathbf{x}^{n+1},\theta,i); \mathbf{H})-\sum_{i\in\theta}\gamma_{i}-\mu.\label{model:pp-lvtw}\tag{PP-L}
\end{gather}

To solve the pricing problem in~\eqref{model:pp-lvtw}, we make the following modifications to the labeling algorithm described in Section~\ref{sec:pricing}.
First, when extending a label $L_\theta$ along an arc $(i,j)$, we generate a new path $\theta'$ with a reduced cost given by:
\begin{equation}
	\overline{C}_{\theta'} = \overline{C}_{\theta} - c_{i0} + c_{ij} + c_{j0} + h(\mathbf{f}(\mathbf{x}^{n+1},\theta',j); \mathbf{H}) - \gamma_j,
\end{equation}
where we replace the sample averaged penalty in Equation~\eqref{eq:label-ext-cost} by the predicted penalty.

Second, we cannot define the earliest service start time $\tau_\theta$ as in Equation~\eqref{eq:label-tau} since we do not assume a set of scenarios to be available under the penalty-based approximation model. Instead, we rely on the following assumption to compute $\tau_\theta$:

\begin{assumption}[Earliest Arrival Times]\label{as:earliest}
Given a route $\theta$, it holds that
\begin{equation}\nonumber
	a_{\theta}(i;\mathbf{\tilde{t}})\geq\min_{k\in\{1,\ldots,n\}}a_{\theta}(i;\mathbf{t}^{k})\qquad\qquad\qquad\forall i\in\theta.
\end{equation}
\end{assumption}
Assumption~\ref{as:earliest} states that, for a given route $\theta$, any realization of the vector of travel times will lead to an arrival time at customer~$i\in\theta$ that is not earlier than the most optimistic arrival time that one would observe given the historical travel times $\{\mathbf{t}^k\}_{k\in\{1,\dots,n\}}$ from the training data set~$\mathbf{T}$. In practice, there often exists periods in which highways have no traffic and cars can travel at free flow, e.g., during quiet night hours. We argue that Assumption~\ref{as:earliest} is reasonable if the historical data contains travel time observations recorded at such periods with low traffic. Even if the historical data does not contain such observations, our assumption is not limiting as one can always augment the data set with artificial scenarios by computing free-flow travel times for each arc. Given the above assumption, we define the earliest service start time:
\begin{equation}
	\tau_{\theta} = \min_{k\in\{1,\dots,n\}} \left\{s_{\theta} (i, \mathbf{t}^k) \right\},
\end{equation}
which is independent of the penalty prediction model $h(\cdot)$.
We run the pricing algorithm from Section~\ref{sec:pricing} with the above modifications to solve~\eqref{model:pp-lvtw}.

\newcommand{\Savg}{\mathcal{S}_{\textsf{avg}}}
\newcommand{\Spto}{\mathcal{S}_{\textsf{pto}}}
\newcommand{\Ssaa}{\mathcal{S}_{\textsf{saa}}}
\newcommand{\Sfbsaa}{\mathcal{S}_{\textsf{fb-saa}}}
\newcommand{\Otest}{\mathcal{T}_{\textsf{test}}}
\newcommand{\Ptest}{\mathcal{X}}
\newcommand{\xtest}{\mathbf{x}'}
\newcommand{\ntest}{n_{\textsf{test}}}

\section{Design of Experiments}\label{sec:design-exp}

The goals of our experiments are to (\textit{i}) compare different data-driven methods for the~\ref{model:vtw}, based on the average cost calculated on a test data set containing travel times and features, (\textit{ii})~analyze the value of incorporating features in the \gls{VRPTW}, and (\textit{iii}) investigate the performance of the proposed methods under different generative models for travel times and features.

We implemented the \gls{BPC} method in \texttt{C++}, using CPLEX as the underlying linear programming solver. All experiments were conducted on the Narval computing cluster from the Digital Research Alliance of Canada, Canada's national high-performance computing system. We set a time limit of 5 hours per run. We provide the source code and data to reproduce our experiments at [to be disclosed after peer-review].


\subsection{Instances}\label{sec:genmodel}

We base our experiments on the instances of~\cite{Solomon_1987}, a standard benchmark for the \gls{VRPTW}. Each instance describes a graph ${\mathcal{G}=(\mathcal{V}, \mathcal{A})}$, a vehicle capacity~$Q$, demands~$q_i$ and time windows~$[e_i, \ell_i]$ of each node $i\in\mathcal{V}$, with the demand for the depot being equal to zero. We define the deterministic cost $c_{ij}$ to be the Euclidean distance between each pair of customers~${(i,j)\in \mathcal{A}}$, and assume a quadratic penalty function, i.e., $\pi(u)=u^2$ for $u\geq0$.

We adopted 29 instances from problem sets R1, C1, and RC1, which have short time windows, allowing only a few customers per route.
Problem set~R1 contains randomly generated geographical data, C1 contains clustered data, and RC1 has a mix of random and clustered structures.


For each instance, we generate synthetic historical data $\mathbf{X}\in\mathbb{R}^{n{\times}p}$ and $\mathbf{T}\in\mathbb{R}^{n{\times}|\mathcal{A}|}$ with $n=100$ training travel time scenarios and $p=10$ features. 
We selected the values for $n$ and $p$ reflecting, e.g., a typical \gls{VRP} application in city logistics where one might have daily observations of features and average travel times for a certain period of the day. For each $k \in \{1, \dots, n\}$, we sample a feature vector from a $p$-variate distribution $\mathbf{x}^k \sim \mathbb{X}$ and we sample travel times from an \mbox{$|\mathcal{A}|$-variate} distribution \mbox{$\mathbf{t}^k \sim \mathbb{T}(\mathbf{x}^k)$} conditioned on the observed feature vector $\mathbf{x}^k$. We consider three different generative models, which specify the distributions $\mathbb{X}$ and $\mathbb{T}$.

\vspace*{0.3cm}
\noindent\textbf{Linear model.} For each arc $(i,j)\in\mathcal{A}$, the deterministic cost $c_{ij}$ defines a nominal travel time~$\underline{t}_{ij}$ corresponding to the free-flow travel time of the arc. We define the stochastic travel times as the nominal travel times $\mathbf{\underline{t}} = [\underline{t}_{ij}]_{(i,j)\in\mathcal{A}}$ plus a random noise term which depends linearly on the features:
\begin{equation}
	\mathbf{\tilde{t}}_{\text{linear}}(\mathbf{\tilde{x}}) = \mathbf{\underline{t}} + \mathbf{B}^{\top}\mathbf{\tilde{x}}+\bm{\tilde{\varepsilon}}
\end{equation}
where $\mathbf{B}$ is a $p \times |\mathcal{A}|$ matrix whose columns are given by the vectors $\mathbf{b}_{ij}\in\mathbb{R}^p$ for $(i,j)\in\mathcal{A}$. 
We sample values in~$\mathbf{b}_{ij}$ from a uniform distribution with support ranging from $1\%$ to $20\%$ of the corresponding nominal travel time~$\underline{t_{ij}}= c_{ij}$. Moreover, we assume that features are binary, ${\mathbf{x} \in \{0,1\}^p}$, corresponding to categorical data, e.g., day of the week, holidays, or roadworks. 
The noise term~$\bm{\tilde{\varepsilon}}$ follows a multivariate normal distribution with zero mean, and covariance matrix generated according to the method of~\cite{Rostami_2021}, such that noise values at different arcs are correlated. Lastly, we assume that travel times on each arc must be greater than or equal to the nominal travel time, and we truncate travel times whenever necessary.

\vspace*{0.3cm}
\noindent\textbf{Exponential model.} For each arc $(i,j)\in\mathcal{A}$, we consider that features are related to the travel times via an exponential function:
\begin{equation}
\tilde{t}_{ij} = \underline{t_{ij}} + 0.2 \,\, \underline{t_{ij}} \, \exp{(2 \, \mathbf{b}_{ij}^{\top} \, \mathbf{\tilde{x}})} + \tilde{\varepsilon}_{ij}
\end{equation}
where features now follow a uniform distribution between 0 and 1. We generate the parameter vectors $\mathbf{b}_{ij}\in\mathbb{R}^p$ by sampling each element from a uniform distribution between 0.1 and 0.3, and we multiply each element by~$-1$ with a probability of 0.2. 
Due to the exponential travel times, having normally distributed $\tilde{\varepsilon}_{ij}$ does not provide sufficient noise. Therefore, we assume that $\tilde{\varepsilon}_{ij}$ follows a log-normal distribution with zero mean and standard deviation $\sigma_{\varepsilon} = 1$.

\vspace*{0.3cm}
\noindent\textbf{Sigmoidal model.} For each arc $(i,j)\in\mathcal{A}$, we generate travel times:
\begin{equation}
\tilde{t}_{ij} = \underline{t_{ij}} + \, \underline{t_{ij}} \, \sigma{\Big(32 (\, \tfrac{1}{2} \mathbf{b}_{ij}^{\top} \, \mathbbm{1} - \mathbf{b}_{ij}^{\top} \, \mathbf{\tilde{x}} )\Big)} + \tilde{\varepsilon}_{ij}
\end{equation}
where $\sigma(x) = 1 / (1+e^{-x})$ is the sigmoid function. Features follow a uniform distribution between~0 and~1.
The noise term $\tilde{\varepsilon}_{ij}$ follows a log-normal distribution with mean 0 and standard deviation $\sigma_{\varepsilon} = 1.2$. We generate the parameter vectors~$\mathbf{b}_{ij}$ by sampling each element from a uniform distribution between 0.3 and 0.8, and we multiply each element by $-1$ with a probability of 0.2. Due to its characteristic shape, we can interpret the sigmoidal model as representing two feature-dependent \textit{states of traffic}, e.g., a congested and a non-congested state.


\subsection{Prescriptive Metrics}

We use the term \textit{prescription} to refer to any function $\mathbf{\hat{z}}(\mathbf{x})$ that provides a decision given the feature observation $\mathbf{\tilde{x}} = \mathbf{x}$.
We evaluate the performance of a given prescription $\mathbf{\hat{z}}(\cdot)$ by its expected cost under the true joint distribution of travel times and features:
\begin{equation}
R(\mathbf{\hat{z}}) = \mathbb{E}_{\mathbf{\tilde{x}}\sim\mathbb{X}}\Big[ \mathbb{E}_{\mathbf{\tilde{t}}\sim\mathbb{T}} \big[ f_{\Theta} (\mathbf{\hat{z}}(\mathbf{x}), \mathbf{\tilde{t}}) \,\big|\, \mathbf{\tilde{x}} = \mathbf{x} \big] \Big]
\end{equation}
In practice, calculating the above expectation is often intractable. We therefore estimate $R(\mathbf{\hat{z}})$ using a test data set, which we construct by sampling a set $\mathcal{X}=\{\mathbf{x}^k\sim\mathbb{X}\}_{k=1,\dots,n_{\mathcal{X}}}$ of features and, for each feature vector $\mathbf{x}\in\mathcal{X}$, a set $\mathcal{T}(\mathbf{x})=\{\mathbf{t}^k \sim \mathbb{T}(\mathbf{x}) \}_{ k=1,\dots,n_{\mathcal{T}}}$ of travel times from the corresponding generative model. The empirical \textit{test cost} $\hat{R}(\mathbf{\hat{z}})$ of a prescription $\mathbf{\hat{z}}(\cdot)$ is an approximation of~$R(\mathbf{\hat{z}})$ which is given by:
\begin{equation}
\hat{R}(\mathbf{\hat{z}}) = \frac{1}{n_{\mathcal{X}} \cdot n_{\mathcal{T}}} \sum_{\mathbf{x}\in\mathcal{X}} \sum_{\mathbf{t}\in\mathcal{T}(\mathbf{x})} f_{\Theta} \left(\mathbf{\hat{z}}(\mathbf{x}), \mathbf{t}\right). \label{eq:test-cost}
\end{equation}
In Section~\ref{sec:results}, we provide test cost results for prescriptions based on different models, where we replace $\mathbf{\hat{z}}$ in Equation~\eqref{eq:test-cost} by the solution to the corresponding model. 

\subsection{Full-information benchmarks}\label{sec:full-benchmarks}

We compare our models against benchmark solutions that rely on knowledge of travel times and feature distributions. The following benchmark solutions are impractical in a real-world setting, as the decision-maker does not know the underlying distributions. However, these benchmark solutions provide us with a relative measure of how well the proposed practical models perform.

\vspace*{0.5cm}
\noindent\textbf{Full-information solution.} Based on the test data set, we can approximate the full-information solution of the~\ref{model:vtw}, given a feature vector $\mathbf{x}\in\mathcal{X}$, as:
\begin{equation}
\mathbf{\hat{z}}^*_{\textsc{full}}(\mathbf{x}) = \arg \min_{\mathbf{z}\in\mathcal{Z}_{\Theta}} \frac{1}{n_{\mathcal{T}}} \sum_{\mathbf{t}\in\mathcal{T}(\mathbf{x})} f_{\Theta} (\mathbf{z}, \mathbf{t}),
\end{equation}
with optimal objective value:
\begin{equation}
\hat{\upsilon}^*_{\textsc{full}}(\mathbf{x}) = \min_{\mathbf{z}\in\mathcal{Z}_{\Theta}} \frac{1}{n_{\mathcal{T}}} \sum_{\mathbf{t}\in\mathcal{T}(\mathbf{x})} f_{\Theta} (\mathbf{z}, \mathbf{t}).
\end{equation}
Given the prescription $\mathbf{\hat{z}}^*_{\textsc{full}}$, the empirical test cost is given by:
\begin{align}
\hat{R}_{\textsc{full}} &= \hat{R}(\mathbf{\hat{z}}^*_{\textsc{full}}) = \frac{1}{n_{\mathcal{X}} \cdot n_{\mathcal{T}}} \sum_{\mathbf{x}\in\mathcal{X}} \sum_{\mathbf{t}\in\mathcal{T}(\mathbf{x})} f_{\Theta} \left(\mathbf{\hat{z}}^*_{\textsc{full}}(\mathbf{x}), \mathbf{t}\right) = \frac{1}{m_{\mathcal{X}}} \sum_{\mathbf{x}\in\mathcal{X}} \hat{\upsilon}^*_{\textsc{full}}(\mathbf{x}),
\end{align}
which provides a lower bound for the empirical test cost of any model. With this definition, we calculate the full-information percentage gap of a prescription $\mathbf{\hat{z}}$ as: 
\begin{equation}
\rho(\mathbf{\hat{z}}) = \frac{\hat{R}(\mathbf{\hat{z}}) - \hat{R}_{\textsc{full}}}{\hat{R}_{\textsc{full}}}.
\end{equation}

\vspace{0.5cm}
\noindent\textbf{Predict with full information, then optimize.}
The performance of a prescription under the \gls{PTO} framework depends on the choice of predictive model $\mathbf{g}(\cdot)$ providing travel time predictions $\mathbf{\hat{t}}=\mathbf{g}(\mathbf{x}; \bm{\varphi})$. 
For the purpose of benchmarking, we define the PTO-F problem, which assumes that the predictive model perfectly predicts the expected travel times given observed features:
\begin{equation}
\mathbf{z}^*_{\textsc{pto-f}}(\mathbf{x}) = \arg \min_{\mathbf{z}\in\mathcal{Z}_{\Theta}} f_{\Theta} \left( \mathbf{z}, \mathbf{\overline{t}} \right) \quad \text{ with } \, \mathbf{\overline{t}} = \mathbb{E} \left[ \mathbf{\tilde{t}} \,\middle|\, \mathbf{\tilde{x}} = \mathbf{x} \right].
\end{equation}
A solution for the PTO-F problem requires knowledge of the joint distribution of travel times and features. We can approximate the PTO-F solution using the test data set:
\begin{equation}
\mathbf{\hat{z}}^*_{\textsc{pto-f}}(\mathbf{x}) = \arg \min_{\mathbf{z}\in\mathcal{Z}_{\Theta}} f_{\Theta} \left( \mathbf{z}, \mathbf{\hat{t}} \right) \quad \text{ with } \, \mathbf{\hat{t}} = \frac{1}{n_{\mathcal{T}}} \sum_{\mathbf{t}\in\mathcal{T}(\mathbf{x})} \mathbf{t}.
\end{equation}
We note that although PTO-F often performs better than \gls{PTO} with practical prediction models, \mbox{PTO-F} does not necessarily provide a lower bound for \gls{PTO} regarding test cost since neither \gls{PTO} nor PTO-F account for the structure of the downstream optimization problem. Specifically, the true conditional expected travel times do not necessarily correspond to the travel times leading to minimum cost.

\subsection{Prescriptive methods}\label{sec:dd-methods}

We compare the ten prescriptive methods summarized in Table \ref{tab:expmodels}. The top eight methods are derived from the data-driven prescriptive models of Section~\ref{sec:dd-prescriptive-models} and correspond to practical methods for solving the~\ref{model:vtw}. The bottom two prescriptive methods are full-information benchmarks, included for comparison, and cannot be applied in practice as they require knowledge of the true travel times distribution (see Section~\ref{sec:full-benchmarks}).

D-avg is the point-based model under average travel times based on historical data. PTO-OLS is the point-based model under the \gls{PTO} framework, where the coefficients $\bm{\varphi}\in\mathbb{R}^{|\mathcal{A}|{\times}p}$ of the prediction model are trained by solving a multiple least squares regression on the historical data. PTO-$k$NN is the point-based model under the \gls{PTO} framework, where we used a $k$-nearest neighbor regression model to predict the travel times. The \gls{SAA} model assumes that the travel time scenarios are given by the historical travel times. SAA-$k$NN corresponds to the approach of \cite{bertsimas2019predictive}, where we adopted a $k$-nearest neighbors regression model for the weight function. As described in Section~\ref{sec:fbsaa}, \gls{CSAA} generates a set of travel time scenarios~$\Omega$ from a multivariate Gaussian distribution with mean and covariance given by~$\mathbf{\hat{B}}^{\top}\mathbf{x}^{n+1}$ and $\bm{\hat{\Sigma}}$, respectively. 
\gls{RSAA} generates a set of travel time scenarios~$\Omega$ based on the residuals of a linear regression model trained on historical data.
P-NN is the penalty-based approximation model in which the late arrival penalty prediction model is a fully-connected neural network with two hidden layers with 100 neurons each, trained with the Adam optimization algorithm to minimize the squared error loss with an $\ell_2$ regularization term equal to $0.1$.
Note that methods D-avg and \gls{SAA} ignore the features and only consider the historical travel times when searching for a solution for the current period~${n+1}$. 

\begin{table}[t]
\footnotesize
\centering
\caption{Data-driven prescriptive methods (top) and full-information benchmarks (bottom).}
\label{tab:expmodels}
{\linespread{1.1}\selectfont
\begin{tabularx}{\textwidth}{lcl}
\hline
Method & Model & Description \\
\hline
D-avg & \eqref{model:pvtw} & Predicted travel times given by average travel times $\mathbf{\hat{t}}=\sum_{k=1}^{n}\mathbf{t}^{k}/n$\\
PTO-OLS & \eqref{model:pvtw} & Predicted travel times $\mathbf{\hat{t}}=\bm{\varphi}\mathbf{x}^{n+1}$ using OLS regression (predict-then-optimize)\\
PTO-$k$NN & \eqref{model:pvtw} & Predicted travel times $\mathbf{\hat{t}}$ given by $k$-NN regression (predict-then-optimize) \\
\acrshort{SAA} & \eqref{model:svtw} & Travel time scenarios are the set of historical travel times $\Omega=\{\mathbf{t}^{1},\ldots,\mathbf{t}^{n}\}$ with $\alpha^\omega=1/n$\\
SAA-$k$NN & \eqref{model:svtw} & Historical travel times with weights $\alpha^\omega$ given by $k$-NN regression \citep{bertsimas2019predictive}\\
\acrshort{CSAA} & \eqref{model:svtw} & Feature-dependent travel time scenarios (see Section \ref{sec:fbsaa})\\
\acrshort{RSAA} & \eqref{model:svtw} & Feature-dependent travel time scenarios based on linear regression residuals (see Section \ref{sec:rsaa})\\
P-NN & \eqref{model:lvtw} & Penalty prediction model based on fully-connected neural network \\
\hline
PTO-F & \eqref{model:pvtw} & Predict-then-optimize under true conditional expected travel times $\mathbf{\hat{t}} = \mathbb{E}_{\mathbf{t}\sim\mathbb{T}}[\mathbf{t} \,|\, \mathbf{x}^{n+1}]$ \\
Full & \eqref{model:svtw} & Full-information lower bound under the true travel times distribution $\mathbf{t}\sim\mathbb{T}(\mathbf{x}^{n+1})$ \\
\hline
\end{tabularx}}
\end{table}

\section{Computational Results}\label{sec:results}

We start our analysis with a small example illustrating the benefit of incorporating information from feature data in the \gls{VRPTW} formulation, which eases the reader to develop an intuition why feature-based approaches can yield results superior to feature-agnostic approaches. We then report results on larger instances, where we discuss the test cost performance of the different prescriptive methods.

\subsection{Illustrative example}\label{sec:illustrative-ex}

\begin{figure}[t]
	\centering
	\begin{subfigure}[b]{0.4\textwidth}
		\centerline{\includegraphics[width=\linewidth]{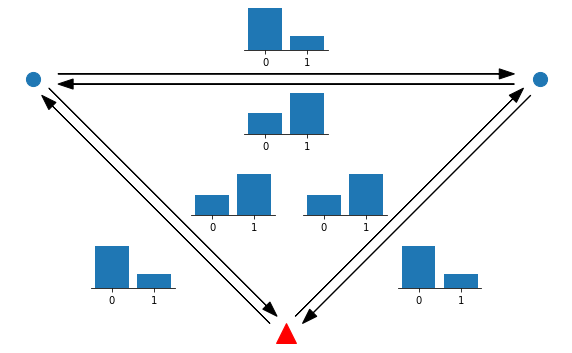}}
		\caption{Travel times as a function of feature value.}
		\label{fig:toy-arc-travel-time-distributions}
	\end{subfigure}
	\hfill
	\begin{subfigure}[b]{0.28\textwidth}
		\centering
		\includegraphics[width=\linewidth]{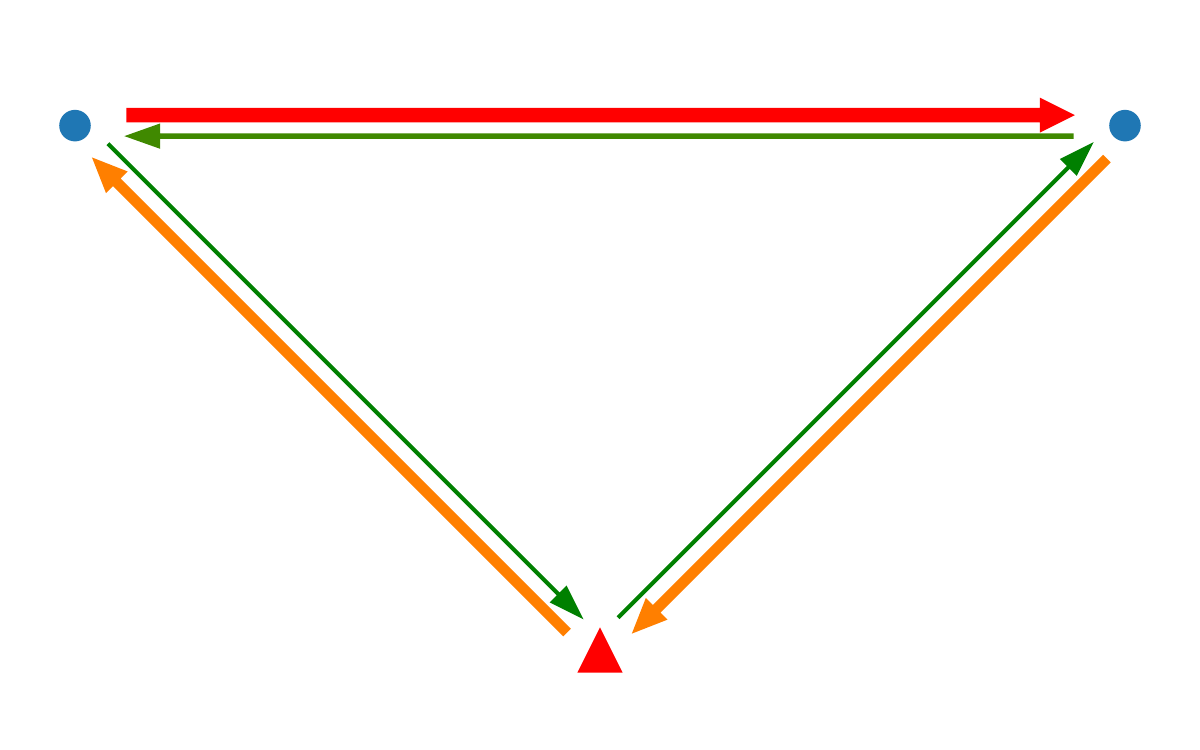}
		\caption{Travel times for $x=0$.}
		\label{fig:toy-example-realizations-x=0}
	\end{subfigure}
	\hfill
	\begin{subfigure}[b]{0.28\textwidth}
		\centering
		\includegraphics[width=\linewidth]{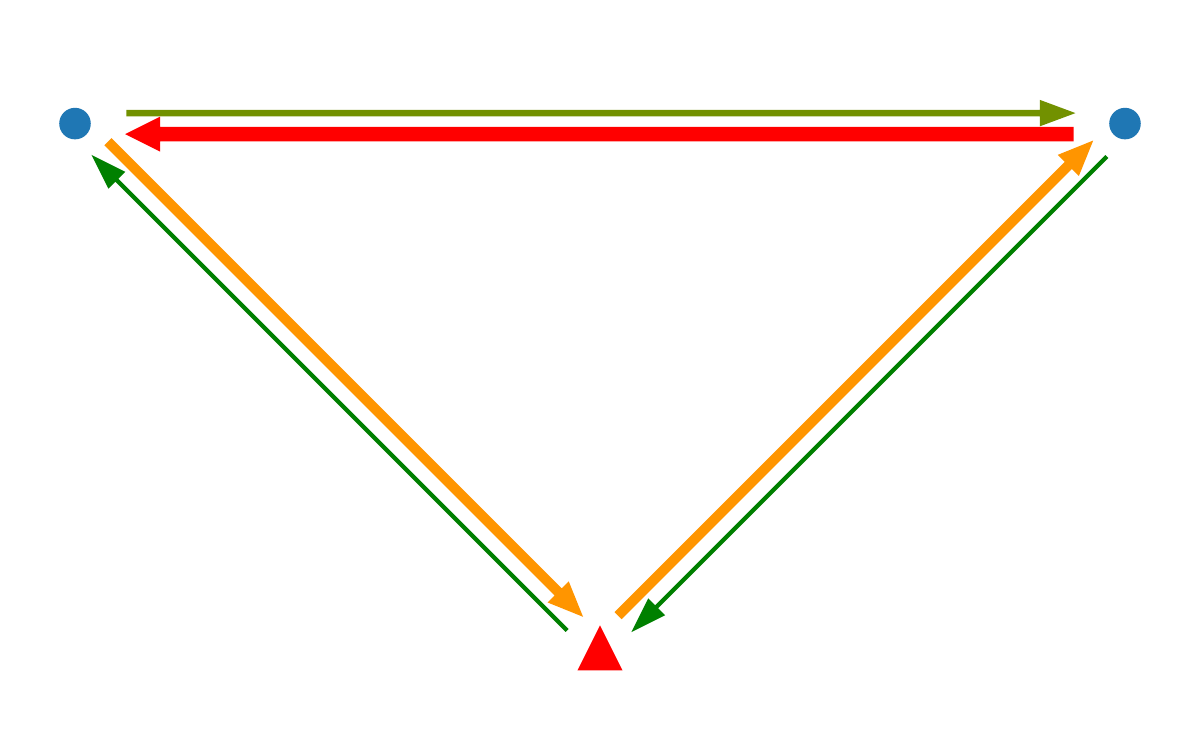}
		\caption{Travel times for $x=1$.}
		\label{fig:toy-example-realizations-x=1}
	\end{subfigure}
	\caption{Joint distribution and different realizations of travel times and feature variable.}
	\label{fig:toy-example-realizations}
\end{figure}

We consider a network with $N=2$ customers and a training data set with $n=2$ samples and $p=1$ binary feature. Figure~\ref{fig:toy-arc-travel-time-distributions} shows the customer locations and arcs connecting each pair of customers. A small bar chart next to each arc shows the corresponding travel times in the y-axis as a function of the feature value in the x-axis. We omitted the values in the y-axis since this illustrative example is not concerned with specific travel time values but rather aims to show the relation between travel times and features. Figure~\ref{fig:toy-example-realizations-x=0} and~\ref{fig:toy-example-realizations-x=1} show the two possible scenarios, representing travel time realizations when the feature value equals $x=0$ and $x=1$, respectively. The color and line thickness of each arc indicate the amount of congestion, with thicker lines representing more congestion, and different colors indicating whether the arc is strongly congested (red), mildly congested (orange), or free from congestion (green). Under the scenario displayed in Figure~\ref{fig:toy-example-realizations-x=0}, a route that starts at the depot and visits customers in a clockwise direction will experience more congestion than a counter-clockwise route. Therefore, if the decision-maker finds herself in a scenario where $x=0$, following the counter-clockwise route is optimal. Figure~\ref{fig:toy-example-realizations-x=1} shows the reverse pattern, i.e., when $x=1$, the counter-clockwise route is more congested than the clockwise route. Existing methods based on \gls{SAA}, which are widely adopted in the field, would provide a clockwise route that is optimal when $x=0$ but not when $x=1$. In contrast to featureless methods, feature-dependent solutions for the CS-VRPTW can provide the optimal route in both cases.

\begin{table}[t]
        \multirow{2}{8cm}{ 
		\begin{subfigure}{0.4\columnwidth}
			\vspace{-4cm}\includegraphics[height=\linewidth]{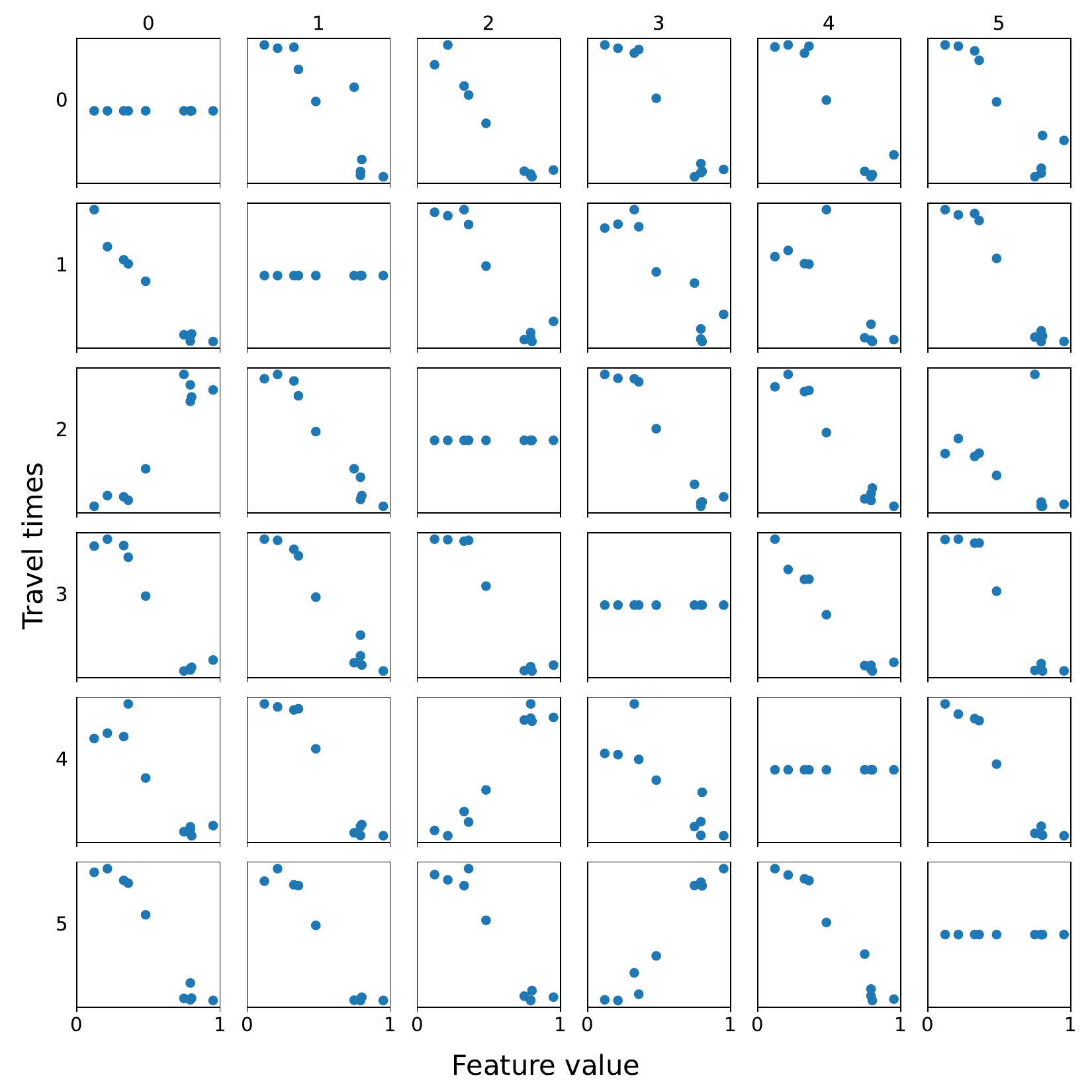}
			\caption{Relation between travel times and features in the training data set for each arc.}\label{fig:toy-instance-tt}
	\end{subfigure}}
	\hspace{-1cm}\begin{subfigure}{0.25\columnwidth}
		\includegraphics[width=1.2\linewidth]{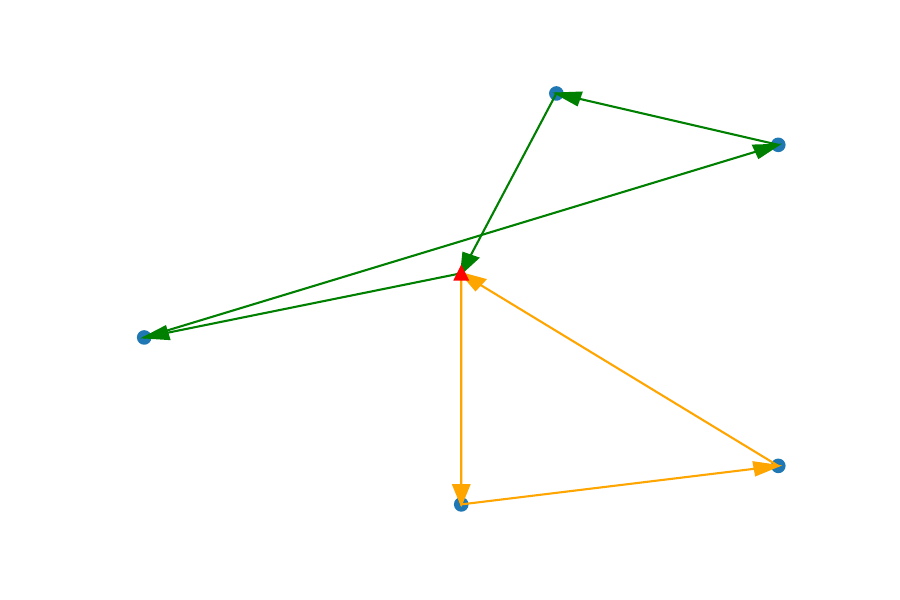}
		\caption{Solution structure A.}\label{fig:toy-example-full-sol-B}
	\end{subfigure}
	\hspace{0.5cm}\begin{subfigure}{0.25\columnwidth}
		\includegraphics[width=1.2\linewidth]{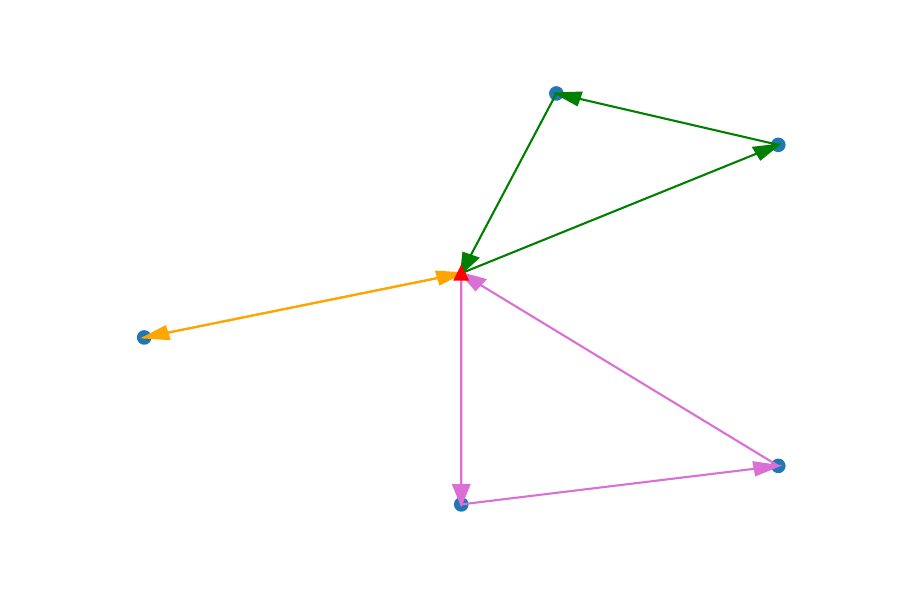}
		\caption{Solution structure B.}\label{fig:toy-example-full-sol-A}  
	\end{subfigure}\\[3em]
	\begin{minipage}{0.4\columnwidth}
	\end{minipage}
	\hfill
	\begin{subfigure}[r]{0.5\columnwidth}
		\subfloat{
			{\linespread{0.6}\selectfont\footnotesize
				\hspace{-0.5cm}\begin{tabularx}{\textwidth}{m{1.6cm} | >{\centering\arraybackslash}L | >{\centering\arraybackslash}L | >{\centering\arraybackslash}L | >{\centering\arraybackslash}L }
					\toprule
					Feature realization & D-avg & SAA & CSAA & Full \\
					\midrule
					$x=0.28$ & \textbf{A} & B & \textbf{A} & \textbf{A} \\
					$x=0.83$ & A & \textbf{B} & \textbf{B} & \textbf{B} \\
					\bottomrule
		\end{tabularx}}}
		\caption*{(d) Optimal solution structures of each method.}
	\end{subfigure}
	\vspace{0.3cm}\captionof{figure}{Data set and solution structures of our illustrative example.} 
	\label{fig:toy-example-full}
\end{table}%

We now extend our previous example and consider a network with $N=5$ customers and a training data set with $n=10$ samples and $p=1$ continuous feature with domain $x\in[0,1]$. To capture different degrees of congestion, we assume sigmoidal travel times (see Section~\ref{sec:genmodel}). In Figure~\ref{fig:toy-instance-tt}, each scatter plot in row $i\in\{0,\dots,5\}$ and column $j\in\{0,\dots,5\}$ shows the travel times and features in the training data set corresponding to arc $(i,j)\in\mathcal{A}$. For a more palatable exposition, we normalized the travel times of each arc based on its nominal free-flow travel time. We compare the optimal solutions of two featureless prescriptive methods, i.e., D-avg and \gls{SAA}, against \gls{CSAA} and full-information solutions. Figures~\ref{fig:toy-example-full-sol-A} and \ref{fig:toy-example-full-sol-B} show the solution structures that emerge in our example. Note how the solution structures are fundamentally different from each other. In particular, Solution A requires three vehicles, while Solution B requires only two vehicles. In Figure~\ref{fig:toy-example-full}d, we show the solution structures obtained by each method under different realizations of the feature variable. When~$x=0.28$, the featureless point-based approximation (D-avg) retrieves the full-information solution, but it fails to do so when~$x=0.83$. We see the opposite behavior for the featureless \gls{SAA} method, i.e., it fails for~${x=0.28}$ but can retrieve the full-information solution when $x=0.83$. In contrast to the featureless approaches, \gls{CSAA} provides feature-dependent solutions that match the full-information solution in both cases. This small example shows that ignoring information from feature data can lead to suboptimal solutions, underlying the benefit of the proposed CS-VRPTW formulation.

\subsection{Results}

We analyze the suitability of the different prescriptive methods in solving the~\ref{model:vtw}. Specifically, we investigate the performance of each method in terms of their test cost~$\hat{R}$ and full-information percentage gaps~$\rho$. To provide a meaningful comparison of the prescriptive methods, we only consider instances for which all methods could find a near-optimal solution with an optimality gap of at most~$1\%$. We restrict the discussion in the subsequent analyses to instances with 25 and 50 customers, for which we obtained a reasonably large set of near-optimal solutions. We provide detailed results for instances with 25 and 50 customers in~\ref{sec:results-25} and \ref{sec:results-50}, respectively. We focus on larger instances with 75 and 100 customers in Section~\ref{sec:scalability}, where we study the scalability of the \gls{BPC} algorithm. As we later discuss, the P-NN method has longer run times and could not solve many instances with 50~customers. Therefore, we show results for P-NN only for instances with 25 customers.

\vspace*{0.5cm}
\noindent\textbf{Average test cost performance.} Table~\ref{res-ttmodel-instance_type} summarizes the test cost performance of all methods, showing the full-information percentage gap of test costs averaged over all instances with a common number of customers $N$, generative model, and instance type. In the last column, we show the absolute test cost values of the full-information benchmark. Among the data-driven prescriptive methods, we highlight the lowest values on each row with boldface numbers.

\begin{table}[htbp]
\footnotesize
\centering
\caption{Overview of average test costs for different instance types and generative models.}
\label{res-ttmodel-instance_type}
\begin{tabularx}{\textwidth}{>{\centering\arraybackslash}m{0.5cm}|m{0.8cm}|m{0.7cm}|LLLLLLLL|LL}
\toprule
 $N$ &  Gen. model & Inst. type & D-avg &  SAA & PTO-OLS & PTO-$k$NN &   SAA-$k$NN & CSAA & RSAA & P-NN & PTO-F &  Full (Abs.) \\
\midrule
    \multirow{9}{*}{25} &  \multirow{3}{*}{Lin.} & R &  23.42 &   3.38 &   0.97 &  11.00 &   4.25 &   \textbf{0.06} &   0.11 &    0.59 &   0.99 & 483.19 \\
    & & C & \textbf{0.00} & \textbf{0.00} & \textbf{0.00} & \textbf{0.00} &   0.48 & \textbf{0.00} & \textbf{0.00} &    0.14 &   0.00 & 192.66 \\
    & & RC & 17.32 &   6.00 &   2.97 &  16.48 &   3.14 &   0.42 &  \textbf{0.38} &    2.56 &   2.11 & 374.16 \\
    \cline{2-13}
    &  \multirow{3}{*}{Exp.} & R &  65.10 &  15.07 &   3.71 &  15.73 &   6.37 &   2.24 &   \textbf{2.07} &    2.54 &   2.18 & 624.45 \\
    & & C & 44.37 &   3.98 &  12.67 &  42.28 &   8.52 &   2.82 &   3.47 & \textbf{2.16} &   7.94 & 220.52 \\
    & & RC & 99.82 &  24.70 &  13.24 &  34.30 &  10.53 &  \textbf{7.75} &   9.58 &   10.52 &   6.03 & 514.96 \\
    \cline{2-13}
    &  \multirow{3}{*}{Sig.} & R &  97.60 &  19.10 &  39.06 &  58.83 &  18.95 &  13.70 &  \textbf{12.89} &   23.03 &  16.96 & 612.18 \\
    & & C & 85.70 & \textbf{30.37} &  74.45 &  72.38 &  45.89 &  41.00 &  42.48 &   47.35 &  81.26 & 231.30 \\
    & & RC & 233.19 &  \textbf{31.12} &  89.02 & 120.53 &  34.35 &  32.77 &  37.99 &   52.42 &  49.80 & 516.79 \\
    \hline
   \multirow{9}{*}{50} & \multirow{3}{*}{Lin.} & R & 23.49 & 4.14 & 2.06 & 14.20 & 3.47 & 0.38 & \textbf{0.31} & - & 1.33 & 917.0 \\
   & & C & 1.14 & 5.22 & 0.49 & 1.14 & 0.89 & \textbf{0.16} & \textbf{0.16} & - & 0.49 & 369.6 \\
   & & RC & 55.09 & 5.76 & 4.87 & 23.38 & 7.84 & 0.71 & \textbf{0.64} & - & 3.69 & 896.9 \\
   \cline{2-13}
   & \multirow{3}{*}{Exp.} & R & 49.12 & 13.31 & 6.96 & 13.75 & 6.27 & \textbf{3.67} & 4.34 & - & 3.76 & 1171.4 \\
   & & C & 45.44 & 10.37 & 5.25 & 30.75 & 3.42 & 2.89 & \textbf{2.36} & - & 5.21 & 432.2 \\
   & & RC & 82.81 & 34.14 & 22.35 & 25.20 & 28.06 & \textbf{6.19} & 9.63 & - & 4.80 & 1274.0 \\
   \cline{2-13}
   & \multirow{3}{*}{Sig.} & R & 191.73 & 34.13 & 67.16 & 95.00 & 32.86 & \textbf{24.37} & 26.51 & - & 27.82 & 1229.0 \\
   & & C & 105.19 & \textbf{23.49} & 62.00 & 67.45 & 28.58 & 34.54 & 29.78 & - & 46.55 & 451.3 \\
   & & RC & 303.64 & 49.66 & 101.90 & 157.20 & 54.73 & \textbf{44.97} & 48.50 & - & 35.49 & 1413.9 \\
\bottomrule
\end{tabularx}
\end{table}

In most settings, \gls{CSAA} and \gls{RSAA} achieve the lowest test costs among the data-driven prescriptive methods. In particular, they outperform the PTO-F benchmark, even though PTO-F has an unfair advantage of having access to the full distribution. As discussed in previous works (cf.~\citealt{bertsimas2019predictive,Elmachtoub_2021}), travel time predictions minimizing a least squares deviation may not lead to optimal decisions since they do not account for the structure of the optimization objective. In line with previous works, our results show that PTO-F often finds suboptimal decisions despite using the true expected travel times under full distributional information.

\begin{figure}[t]
	\centering
	\includegraphics[width=1.01\textwidth]{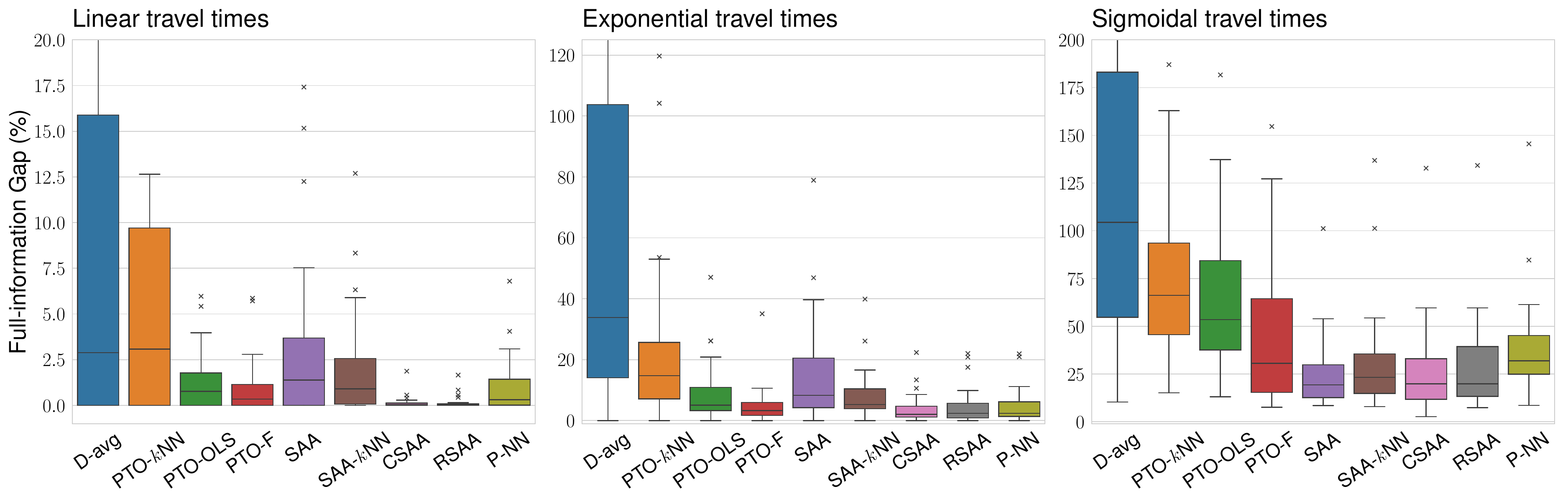} 
	\caption{Distribution of test costs for instances with 25 customers from different generative models.}
	\label{fig:boxplots-test_cost-25}
\end{figure}

\vspace*{0.5cm}
\noindent\textbf{Analysis of test cost distributions.} We focus on the distribution of test costs for instances with 25 customers. 
Figure~\ref{fig:boxplots-test_cost-25} presents full-information gaps of each method, summarized as boxplots with whiskers that extend to 1.5 times the
interquartile range. Points outside this range are marked as outliers and noted with an~``$\times$''.

We first note that, as expected, \gls{CSAA} and \gls{RSAA} have the smallest gaps for instances with linear travel times, as the data in this case conforms with the model assumptions. Second, we observe that P-NN provides competitive solutions in general and, especially for instances with exponential travel times, often outperforms all other methods. In particular, P-NN often improves upon \gls{PTO}-based methods, highlighting the advantage of directly predicting the late arrival penalty rather than relying on travel time predictions. Still, \gls{CSAA} and \gls{RSAA} present equal or even superior performance in this non-linear setting. Third, for instances with sigmoidal travel times, we observe that \gls{SAA}-based methods, i.e., \gls{SAA}-$k$NN, \gls{CSAA}, \gls{RSAA}, and the classical \gls{SAA}, have the smallest full-information gaps. In this highly non-linear setting, we benefit from using scenarios that better capture travel time variability, as illustrated in our example in Section~\ref{sec:illustrative-ex}. In general, we conclude that simpler methods that extend \gls{SAA} using feature-dependent scenarios are the most promising, providing solutions that are closest to the full-information benchmark. Although it might be surprising that \gls{CSAA} and \gls{RSAA} can outperform the more sophisticated P-NN method, we argue that predicting late arrival penalties is not straightforward because the training data set must reflect the relation between the different combinatorial structures, encoded in the projected features, and the resulting penalty. The proposed penalty prediction model is not always capable of learning all the combinatorial intricacies of the VRP problem.

\begin{figure}[t]
	\centering
	\begin{subfigure}{.34\textwidth}
		\centering
		\hspace{-0.5cm}\includegraphics[width=\textwidth]{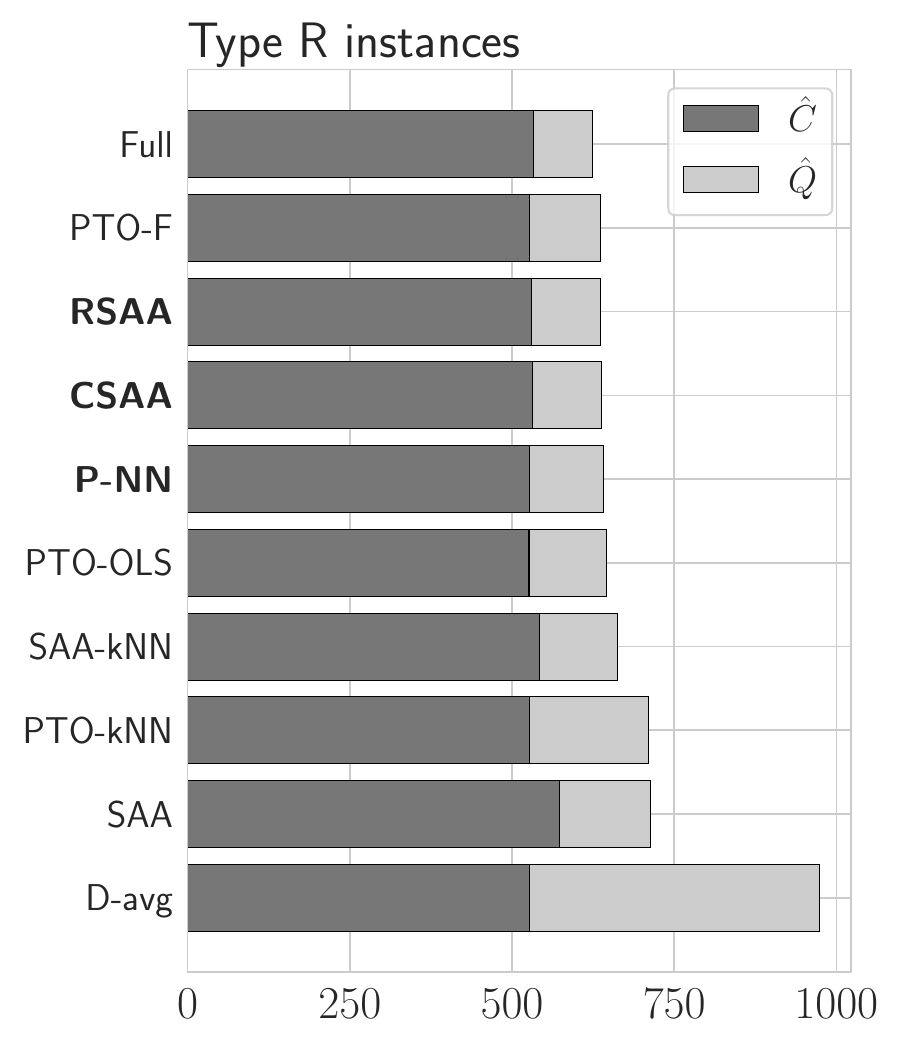}
	\end{subfigure}%
	\begin{subfigure}{.34\textwidth}
		\centering
		\hspace{-0.5cm}\includegraphics[width=\textwidth]{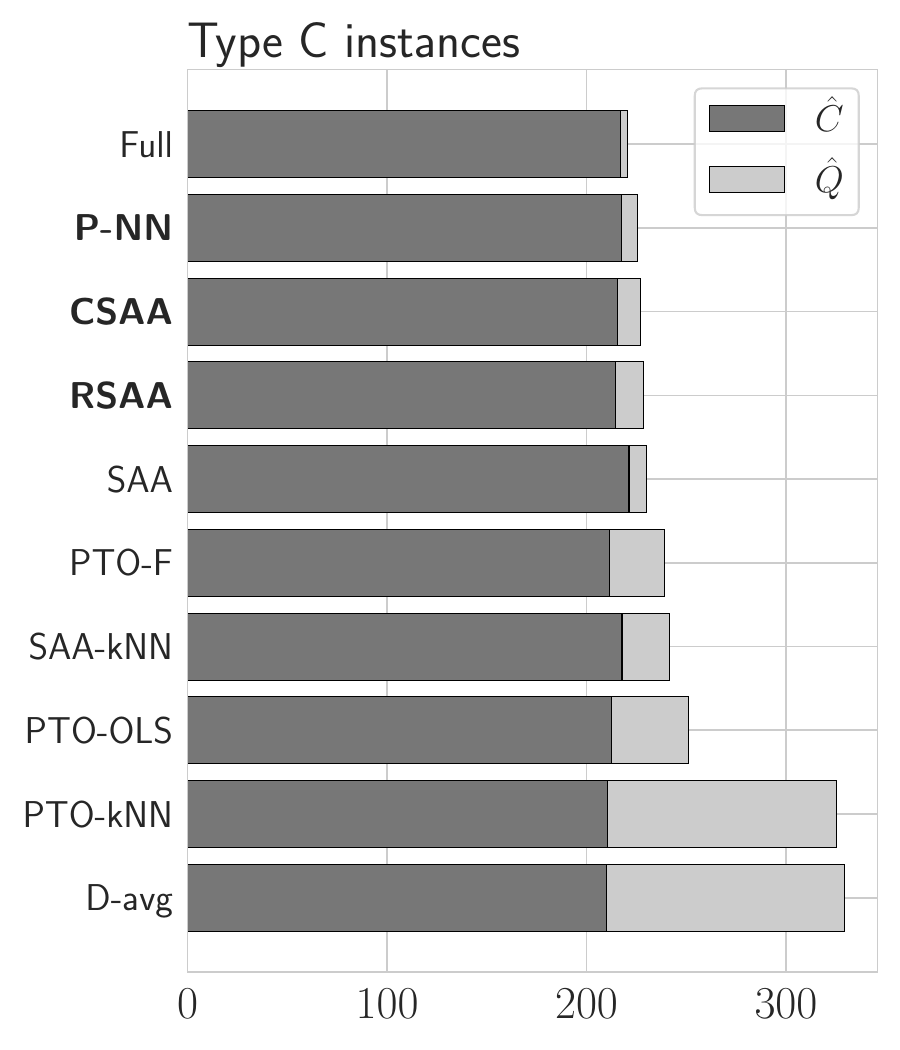}
	\end{subfigure}%
	\begin{subfigure}{.34\textwidth}
		\centering
		\hspace{-0.5cm}\includegraphics[width=\textwidth]{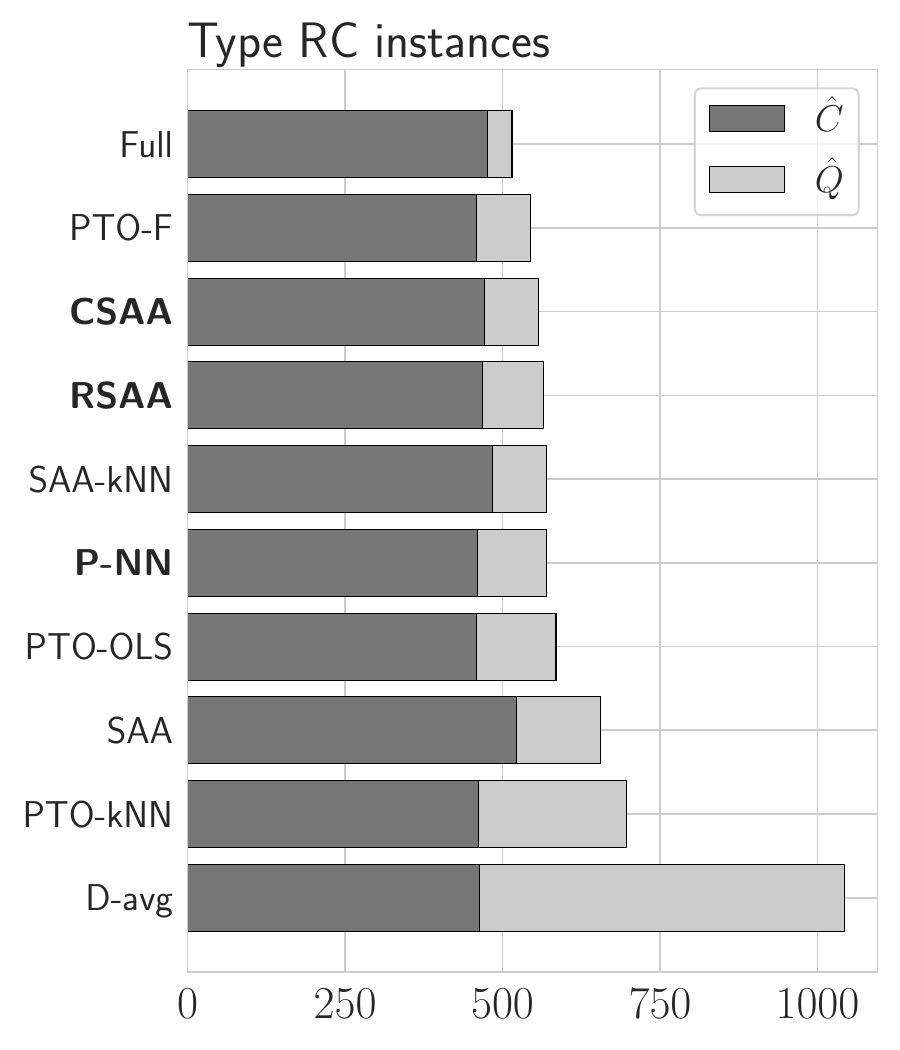}
	\end{subfigure}
	\caption{Average first-stage and second-stage test costs for instances from the exponential generative model.}
	\label{fig:test_cost_penalty}
\end{figure}

\vspace{0.5cm}
\noindent\textbf{Trade-offs between first and second-stage costs.} We analyze the contributions of the first-stage and second-stage costs to the total test cost.
Figure~\ref{fig:test_cost_penalty} displays the average test cost of each method for instances with an exponential generative model, where we decompose the test cost $\hat{R}(\mathbf{\hat{z}})$ of a solution as the sum of the first-stage cost $\hat{C}$ and the second-stage cost $\hat{Q}$:
\begin{align}
	\hat{R}(\mathbf{\hat{z}}) &\,=\, \frac{1}{n_{\mathcal{X}} \cdot n_\mathcal{T}} \sum_{\mathbf{x}\in\mathcal{X}} \sum_{\mathbf{t}\in\mathcal{T}(\mathbf{x})} C\left(\mathbf{\hat{z}}\right) + Q(\mathbf{\hat{z}}, \mathbf{t}) \,=\, \hat{C}(\mathbf{\hat{z}}) + \hat{Q}(\mathbf{\hat{z}}).
\end{align}
The top, middle, and bottom plots show results for instances of types R, C, and RC, respectively. Full-information solutions have first-stage costs that are often larger than those of other methods, showing that it is often beneficial to incur larger first-stage costs and compensate them with comparatively small late arrival penalties. On the contrary, optimizing for the first-stage cost can lead to solutions with large penalties and, consequently, large overall test costs, as we see, e.g., in the middle plot for the D-avg method. Compared to practical \gls{PTO} methods, i.e., PTO-OLS and PTO-$k$NN, our proposed \gls{CSAA} and P-NN methods often present larger first-stage costs but smaller total test costs, bearing more similarity to the results from the full-information benchmark.
	
\subsection{Scaling to larger instances}
\label{sec:scalability}

\begin{figure}[t]
	\centering
	\includegraphics[width=0.9\linewidth]{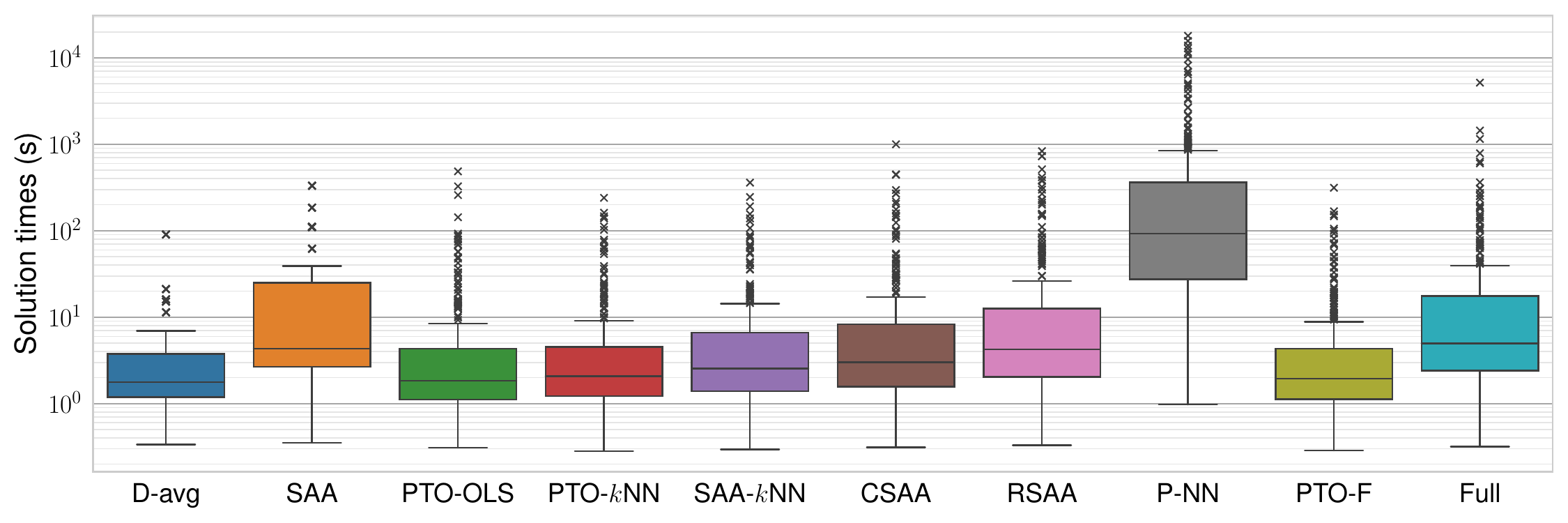} 
	\caption{Solution times of different methods for instances with 25 customers with a linear generative model.}
	\label{fig:boxplots-solution_times-25-linear}
\end{figure}

Figure~\ref{fig:boxplots-solution_times-25-linear} compares the solution times of the different methods. For this plot, we considered all instances with 25 customers and a linear generative model. We observe that P-NN shows the longest run times, followed by \gls{SAA} and the full-information benchmark. Recall that we solve all methods using the \gls{BPC} algorithm described in Section~\ref{sec:algoddvrptw}. Specifically for the P-NN method, we adopt the modified pricing problem described in Section~\ref{sec:algo-penalty-based}. Consequently, with the exception of the P-NN method, the solution algorithms to all data-driven models differ only in the specific travel time scenarios given as input to the \gls{BPC} algorithm.
	
\begin{figure}[t]
	\centering
	\includegraphics[width=0.5\linewidth]{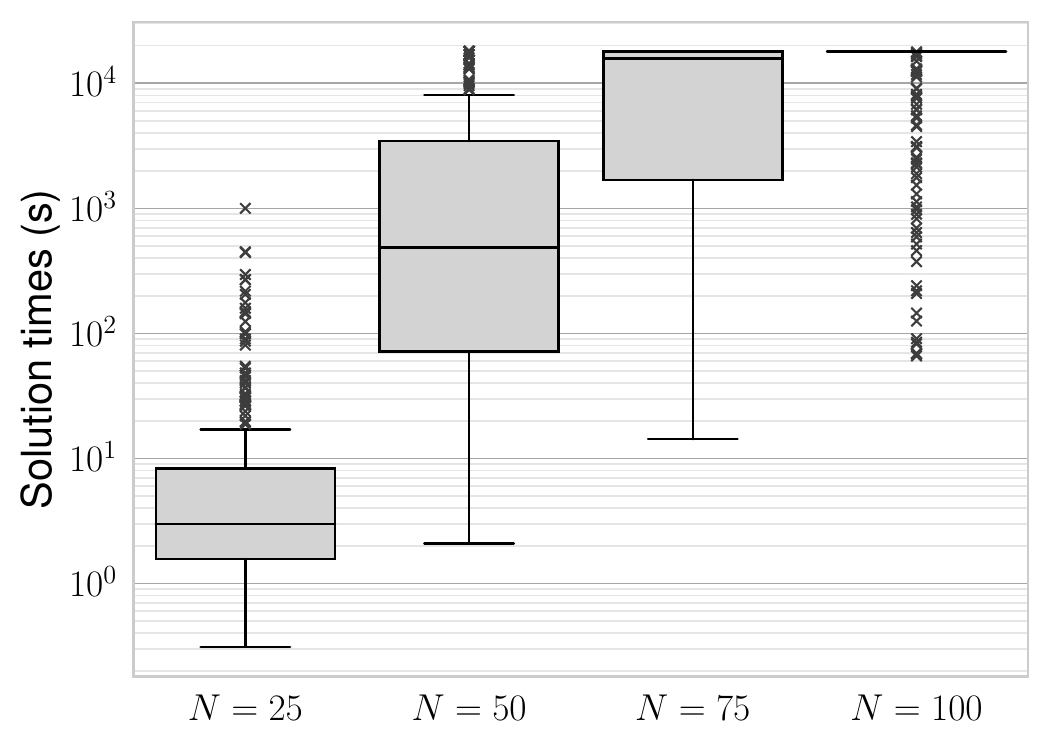} 
	\caption{Solution times of the \gls{CSAA} method for increasing number of customers on instances with a linear generative model.}
	\label{fig:boxplots-solution_times-fbsaa-linear}
\end{figure}

Figure~\ref{fig:boxplots-solution_times-fbsaa-linear} shows how solution times increase as we increase the number of customers. Here, we focus on the \gls{CSAA} method and consider instances with a linear generative model. With~$25$ customers, we can solve most instances in less than 10 seconds. Due to the $\mathcal{NP}$-hardness of the problem, solution times increase drastically, and we observe some runs reaching the 5-hour time limit already with 50 customers. With 100 customers, we can only solve very few instances within the time limit.

\section{Conclusions}\label{sec:conclusion}
	
This work connects the recent literature on contextual optimization with the established research field of \glspl{VRP}.
We introduced a novel formulation incorporating contextual information into the stochastic \gls{VRPTW}. To solve the proposed formulation, we derived several data-driven prescriptive models by ($i$) applying existing contextual optimization methods to our problem setting and ($ii$) proposing novel methods based on conditional \gls{SAA} and penalty-based prediction.
From a computational perspective, solving the \gls{VRPTW} is a challenging task for which many solution techniques exist in the literature. We showed how state-of-the-art techniques can be adapted when contextual information is available, leading to a customized \gls{BPC} algorithm that can solve instances with up to 100 customers. 
We analyzed the out-of-sample cost performance of the data-driven methods. We observed that the penalty-based approximation model that relies on penalty predictions provides competitive solutions, but in general, an \gls{SAA} method based on feature-dependent scenarios yields solutions that are closest to the full-information benchmark.

This work raises several possibilities for future research. First, although we studied various data-driven methods and compared them against different benchmarks, we certainly did not exhaustively explore all possible approaches for solving the \ref{model:vtw}. Future research could investigate variations of the presented methods, e.g., the use of non-linear probabilistic models within the \gls{CSAA} approach, or propose novel models and algorithms. Second, while extending the presented techniques to some \gls{VRP} variants might be straightforward, e.g., with stochastic service times, other problem variants could lead to more interesting research problems. Finally, since this paper focused on modeling aspects and exact solution methods, one open question is how to adapt existing heuristic methods to harness contextual information to efficiently solve larger instances. 

\vspace*{0.5cm}
\noindent\textbf{Acknowledgments} 

This research has been funded by the Deutsche Forschungsgemeinschaft (DFG, German Research Foundation) as part of the research group Advanced Optimization in a Networked Economy (AdONE, GRK2201/277991500). This support is gratefully acknowledged. This research was enabled in part by support provided by Calcul Québec (\url{https://www.calculquebec.ca/}) and the Digital Research Alliance of Canada (\url{https://alliancecan.ca}).


\appendix

\section{Feature Projection Function.}\label{sec:fpf}
The \gls{FPF} is a function $\mathbf{f}:\mathbb{R}^{p}\times\Theta\times\mathcal{V}\setminus\{0\}\mapsto\mathbb{R}^{\bar{p}}$ that generates a \mbox{$\bar{p}$-dimensional} vector of projected features given a feature vector $\mathbf{x}\in\mathbb{R}^p$, a route $\theta\in\Theta$, a node~${i\in\theta}$, and travel time estimates~$\mathbf{\hat{t}}$. The projected features are split into groups, as follows:

\begin{table}[h]
	\footnotesize
	\centering
	\caption{Feature Projection Function: Predictors of the Penalty at Customer $i\in\theta$ Returned by $\mathbf{f}(\mathbf{x},\theta,i;\mathbf{\hat{t}})$\label{tab:feattw}}
	{\begin{tabular}{lcl}
			\toprule Predictor & Group & Description \\
			\midrule
			$x_1, \dots, x_p$ & $(a)$ & Travel time covariates (original features)\\
			\midrule
			$e_i$ & $(b)$ & Start of time window of customer $i\in\theta$\\
			$\ell_i$ & $(b)$ & End of time window of customer $i\in\theta$\\
			$\ell_{\rho(i)}$ & $(b)$ & End of time window of customer $\rho(i)\in\theta$ that precedes $i\in\theta$\\
			$c_{\rho(i),i}$ & $(b)$ & Transportation cost of the arc from $\rho(i)$ to $i$\\
			$\hat{\sigma}_{\rho(i),i}^{2}$ & $(b)$ & Estimated variance in the travel time from $\rho(i)$ to $i$\\
			$k_{i}$ & $(b)$ & Position of customer $i$ along route $\theta$\\
			\midrule
			$a_{\theta}(i;\mathbf{\underline{t}})$ & $(c)$ & Lower bound on arrival time at customer $i\in\theta$ with free-flow travel times $\mathbf{\underline{t}}$\\
			$s_{\theta}(i;\mathbf{\underline{t}})$ & $(c)$ & Lower bound on service start time at customer $i\in\theta$ with free-flow travel times $\mathbf{\underline{t}}$\\
			$(a_{\theta}(i;\mathbf{\underline{t}})-\ell_{i})^{+}$ & $(c)$ & Lower bound on lateness\\
			$\pi(a_{\theta}(i;\mathbf{\underline{t}})-\ell_{i})$ & $(c)$ & Lower bound on penalty\\
			\midrule
			$\hat{t}_{\rho(i),i}$ & $(d)$ & Predicted travel time of the arc from $\rho(i)$ to $i$\\
			$a_{\theta}(i;\mathbf{\hat{t}})$ & $(d)$ & Arrival times at customer $i\in\theta$ given predicted travel times $\mathbf{\hat{t}}$\\
			$s_{\theta}(i;\mathbf{\hat{t}})$ & $(d)$ & Service start time at customer $i\in\theta$ given predicted travel times $\mathbf{\hat{t}}$\\
			$(a_{\theta}(i;\mathbf{\hat{t}})-\ell_{i})^{+}$ & $(d)$ & Lateness at customer $i\in\theta$ given predicted travel times $\mathbf{\hat{t}}$\\
			$\pi(a_{\theta}(i;\mathbf{\hat{t}})-\ell_{i})$ & $(d)$ & Penalty  at customer $i\in\theta$ given predicted travel times $\mathbf{\hat{t}}$\\
			\midrule
			$\xi_{\theta}(i;\hat{\mathbf{t}})$ & $(e)$ & Variability model of service start time at customer $i\in\theta$ (see~\ref{serv-risk})\\
			\bottomrule
	\end{tabular}}
	{}
\end{table}


\subsection{Variability of service start time.}\label{serv-risk}
In order to derive penalty predictors that take into account travel time variability and its impact on the service start time, we introduce a measure of variability of service start time, which we denote as \emph{service start time risk}. Given a route $\theta$ and a customer $i\in\theta$, the distribution of $\max\{e_{i},a_{\theta}(i;\mathbf{\tilde{t}})\}$ (the service start time at customer $i$) is, in general, truncated, because of possible early arrivals and waiting times at customer $i$ and at other customers previously visited. Such a truncation decreases service start time variability; therefore, a covariate for service start time risk should consider both travel time variability and the likelihood of early arrivals along a route.

Let $\bm{\Sigma}=[\sigma_{ij,lm}]_{(i,j),(l,m)\in\mathcal{A}}$ be the travel times covariance, and let $\sigma_{ij}^{2}=\sigma_{ij,ij}$. Given a route ${\theta=(v_{1},\ldots,v_{L})}$, we denote by $\xi_{\theta}(i)$ the service start time risk at customer $i\in\theta$, and consider the following risk propagation model:
\begin{equation}\label{eq:riskm}
	\xi_{\theta}(v_{k})=\begin{cases}
		(1-\mathcal{P}_{\theta}(v_{k}))\sigma_{0v_{k}}^{2},&\text{if }k=1,\\
		(1-\mathcal{P}_{\theta}(v_{k}))(\xi_{\theta}(v_{k-1})+\sigma_{v_{k-1}v_{k}}^{2}+2\sigma_{v_{k-2}v_{k-1},v_{k-1}v_{k}}),&\text{otherwise},
	\end{cases}
\end{equation}
where $\mathcal{P}_{\theta}(i)=\mathbb{P}(a_{\theta}(i;\mathbf{\tilde{t}})<e_{i}\given\mathbf{x}^{n+1})$ is the early arrival probability at customer $i\in\theta$ conditional on observed features $\mathbf{x}^{n+1}$, and $v_{0}=0$.

Model \eqref{eq:riskm} propagates the variabilities of travel time and service start time along route $\theta$ as far as early arrival probabilities are low. When the early arrival probability $\mathcal{P}_{\theta}(i)$ is high, the service start time risk at customer $i$ is low, since service occurs at instant $e_{i}$ with high probability. In this case, the service start time risk at other customers along route $\theta$ following customer $i$ also decreases. Note that the model accounts for travel time correlation between adjacent arcs in the network.

Service start time risk measures $\xi_{\theta}(i)$, $i\in\theta$, cannot be computed directly because the travel time distribution is unknown. Following our distribution-free approach, we estimate $\bm{\Sigma}$ and $\mathcal{P}_{\theta}(i)$ (and hence $\xi_{\theta}(i)$) from data. Let $\bm{\hat{\Sigma}}=[\hat{\sigma}_{ij,lm}]_{(i,j),(l,m)\in\mathcal{A}}$ be the estimated travel times covariance (as described in Section \ref{sec:fbsaa}) and let $\hat{\sigma}_{ij}^{2}=\hat{\sigma}_{ij,ij}$. In the remainder of this section, we discuss how to estimate $\mathcal{P}_{\theta}(i)$ for any route $\theta$. To this end, we let $\mathbf{g}:\mathbb{R}^{p}\mapsto\mathbb{R}^{|\mathcal{A}|}$ be a travel time prediction model. Further, we assume that for each customer $i\in\mathcal{V}\setminus\{0\}$ a set of training routes $\Theta_{i}$ is available, where $i\in\theta$ for all $\theta\in\Theta_{i}$. This is an unrestrictive assumption as these training routes may be arbitrary routes, e.g., generated by solving other \gls{VRP} models. Finally, let $\rho_{\theta}(v_{k})$ be the node that precedes $v_{k}$ in route $\theta=(v_{1},\ldots,v_{L})$, that is, $\rho_{\theta}(v_{k})=0$ if $k=1$, and $\rho_{\theta}(v_{k})=v_{k-1}$ if $k\geq2$.

Given a vector $\mathbf{x}$ of travel time covariates, let $\mathbf{w}_{i,\theta}(\mathbf{x})$ be the vector of early arrival covariates, with components as described in Table~\ref{tab:early_arrival_predictors}. Clearly, if $e_i=0$, then we have $\mathcal{P}_{\theta}(i)=0$. Further, note that if $C_{\theta}>e_i$, then $\mathcal{P}_{\theta}(i)=0$, since we assume that $\tilde{t}_{ij}\geq c_{ij}$. Finally, if $\max_{j\in\theta\setminus\{i\}}\{e_j\}>e_i$, then we have again ${\mathcal{P}_{\theta}(i)=0}$.

\begin{table}[h]
	\centering
	\caption{Components of the early arrival covariates vector $\mathbf{w}_{i,\theta}(\mathbf{x})$ at customer $i\in\theta$ given feature vector $\mathbf{x}$\label{tab:early_arrival_predictors}}
	{\begin{tabular}{ll}
			\hline
			Predictor & Description \\
			\hline
			$x_1, \dots, x_p$ & Travel time covariates\\
			$e_i$ & Opening of time window \\
			$a_{\theta}(i;\mathbf{g}(\mathbf{x}))$ & Estimated arrival time at customer $i\in\theta$\\
			$\hat{\sigma}_{\rho_{\theta}(i)i}^{2}$ & Estimated variance in the travel time from $\rho_{\theta}(i)$ to $i$\\
			$\max_{j\in\theta\setminus\{i\}}\{e_j\}$ & Latest $e_j$ along route $\theta$ before arriving at customer $i$\\
			$C_{\theta}$ & Transportation cost of route $\theta$ up to customer~$i$\\
			\hline
	\end{tabular}}
	{}
\end{table}

Since our predicted quantity is a probability, a sensible learning model is a logistic regression. Let $S(z)=1/(1+\exp(-z))$ be the logistic function, and let $\mathcal{L}_{\text{nll}}(a,b)=-(b\log a+(1-b)\log(1-a))$ be the negative log likelihood loss function. For each customer $i\in\mathcal{V}\setminus\{0\}$, we train the parameters $\hat{\phi}_{i}^{0}$ and $\bm{\hat{\phi}}_{i}\in\mathbb{R}^{p+2}$ of the logit model:
\begin{equation}\nonumber
	\hat{\phi}_{i}^{0},\bm{\hat{\phi}}_{i}=\arg \min_{\phi_{i}^{0}\in\mathbb{R},\bm{\phi}_{i}\in\mathbb{R}^{p+2}}\quad\frac{1}{n|\Theta_{i}|}\sum_{k=1}^{n}\sum_{\theta\in\Theta_{i}}\mathcal{L}_{\text{nll}}\left(S(\phi_{i}^{0}+\bm{\phi}_{i}^{\top}\mathbf{w}_{i,\theta}^{k}),\mathbb{I}(a_{\theta}(i;\mathbf{t}^{k})\leq e_{i})\right)+\lambda\norm{\bm{\phi}_{i}}_{1},
\end{equation}
where we use $\mathbf{w}_{i,\theta}^{k}\coloneqq\mathbf{w}_{i,\theta}(\mathbf{x}^k)$, $\lambda$ is the regularization parameter and $\norm{\bm{\phi}_{i}}_{1}$ is the $\ell_{1}$ norm of $\bm{\phi}_{i}$. Regularization by the $\ell_{1}$ norm leads to sparsity in the model parameters.

Hence, for any route $\theta$ we estimate the early arrival probability at customer $i\in\theta$ by
\begin{equation}\nonumber
	\hat{\mathcal{P}}_{\theta}(i)=S(\hat{\phi}_{i}^{0}+\bm{\hat{\phi}}_{i}^{\top}\mathbf{w}_{i,\theta}^{n+1}).
\end{equation}

\SetKwComment{Comment}{// }{}
\newcommand\mycommfont[1]{\scriptsize\ttfamily\textcolor{blue}{#1}}
\SetCommentSty{mycommfont}

\section{Dynamic programming algorithm for RCSP}\label{appendix-RCSP-bound}

We present a \gls{DP} algorithm for obtaining a \gls{RCSP} bound.  
In Algorithm~\ref{alg:rcsp-bound}, variable $T_1[\delta,i,q]$ stores a lower bound on the reduced cost of extending a route that ends at customer $i$ with remaining capacity $q$, departing from $i$ at time $\delta$ (i.e., the arrival time at customer $i$ is equal to $\delta$). We relax elementarity by allowing routes with cycles but we still remove 2-cycles. Similarly, $T_2[\delta,i,q]$ stores the second best lower bound on the reduced cost. Finally, $N[\delta,i,q]$ stores the customer following $i$ on the route associated with the best lower bound.

\bigskip
\bigskip

\begin{algorithm}[H]
\footnotesize
\linespread{1.4}\selectfont
\DontPrintSemicolon
\caption{Dynamic programming algorithm for RCSP}
\label{alg:rcsp-bound}
\KwResult{matrix $T_1$ of lower bounds on the reduced costs of route extensions}
$\ell_{\max} \gets \max_{i \in \mathcal{V}^+}\{\ell_i\}$\Comment*{latest end of time window among all customers}
$\Delta t \gets \ell_{\max} / 40$ \Comment*{define a time step}
\For{$\delta=0,\Delta t,2 \Delta t, \dots, \ell_{\max}$}{
    $T_1[\delta,i,q]\gets \infty, \quad \text{for } i\in\mathcal{V}^+, \, q=1,\dots, Q$ \Comment*{initialize matrix $T_1$: lower bound on reduced costs}
    $T_1[\delta,0,q]\gets0, \quad \text{for } q=0,\dots, Q$ \Comment*{initialize matrix $T_1$}
    $T_1[\delta,i,0]\gets c_{i0}, \quad \text{for } i\in\mathcal{V}^+$ \Comment*{initialize matrix $B$}
    $T_2[\delta,i,q]\gets \infty, \quad \text{for } i\in\mathcal{V}, \, q=0,\dots, Q$ \Comment*{initialize matrix $T_2$: second best cost}
    $N[\delta,i,q]\gets 0, \quad \text{for } i\in\mathcal{V}, \, q=0,\dots, Q$ \Comment*{initialize matrix $N$: next customer in the route}
    \For{$q=1,\dots,Q$}{
        \For{$i\in\mathcal{V}^+$}{
          $T_1[\delta,i,q] \gets T_1[\delta,i,q-1]$\;
          $T_2[\delta,i,q] \gets T_2[\delta,i,q-1]$\;
          $N[\delta,i,q] \gets N[\delta,i,q-1]$\;
          \For{$j\in\mathcal{V}^+$}{
            \If{$j=i$ or $q_j > q$ or arc $(i,j)$ is forbidden by branching}{
              \Continue \Comment*{does not extend label $L$ to customer $j$} 
            }
            \If{$N[\delta,j,q-q_j]\neq i$}{ 
              $v\gets c_{ij} - \gamma_j + T_1[\delta,j,q-q_j] + \pi(\delta - \ell_j)$ \Comment*{xxx}
            }
            \Else{
              $v\gets c_{ij} - \gamma_j + T_2[\delta,j,q-q_j] + \pi(\delta - \ell_j)$ \Comment*{avoid 2-cycles}
            }
            \If{$v<T_1[i,q]$}{
              $T_2[\delta,i,q] \gets T_1[\delta,i,q]$ \Comment*{move best to second best}
              $T_1[\delta,i,q] \gets v$ \Comment*{set new best}
              $N[\delta,i,q] \gets j$ \Comment*{set next customer}
            }
            \ElseIf{$v<T_2[\delta,i,q]$}{
              $T_2[\delta,i,q] \gets v$ \Comment*{just update second best}
            }
          }
        }
    }
}
\Return $T_1$\;
\end{algorithm}

\section{Proof of completion bounds}\label{sec:proof-completion-bounds}

We provide proofs for the \gls{RCSP} and knapsack bounds in the following.

\vspace{0.2cm}
\textit{Proof of Proposition~\ref{prop:completion-bound-rcsp}.}\hspace*{0.2cm}From the definition of the \gls{RCSP} bound in Equation~\eqref{eq:rcsp-bound-def}, we have:
\begin{align*}
        0 \labelrel\leq{prop-ineq-rcsp-0}\,& \overline{C}_{\theta} + \widehat{T}_{\textsc{RCSP}}(i, Q-q_\theta) \label{proof:completion-bound-rcsp} \\
        \labelrel\leq{prop-ineq-rcsp-1}\,& \overline{C}_\theta - c_{i0} + c_{i u_1} + \pi (\delta_{\theta\oplus u_1} - \ell_{u_1}) - \gamma_{u_1} + c_{u_1 0} + \widehat{T}_{\textsc{RCSP}}(u_1, Q - q_{\theta\oplus u_1}) \nonumber\\
	\labelrel\leq{prop-ineq-rcsp-2}\,& \overline{C}_\theta - c_{i0} + c_{i u_1} + \pi(\delta_{\theta\oplus u_1} - \ell_{u_1}) - \gamma_{u_1} + c_{u_1 0} \nonumber\\
	& \quad - c_{u_1 0} + c_{u_1 u_2} + \pi(\delta_{\theta\oplus u_1 \oplus u_2} - \ell_{u_2}) - \gamma_{u_2} + c_{u_2 0} \nonumber\\ 
	& \quad \dots \nonumber\\
	& \quad - c_{u_{L-1} 0} + c_{u_{L-1} u_L} + \pi(\delta_{\theta\oplus\mathcal{E}} - \ell_{u_L}) - \gamma_{u_L} + c_{u_L 0} + \widehat{T}_{\textsc{RSCP}} (u_L, Q - q_{\theta \oplus \mathcal{E}}) \nonumber\\ 
	\labelrel={prop-ineq-rcsp-3}\,& \overline{C}_\theta - c_{i0} + c_{i u_1} +\sum_{j=2}^{L} c_{u_{j-1} u_j} + c_{u_L 0} + \sum_{j=1}^{L} \Big( \pi(\delta_{\theta\oplus{u_1}\oplus{\dots}\oplus{u_j}} - \ell_{u_j}) - \gamma_{u_j}\Big) + \widehat{T}_{\textsc{RSCP}} (u_L, Q - q_{\theta'}) \nonumber\\
	\labelrel\leq{prop-ineq-rcsp-4}\,& \overline{C}_\theta - c_{i0} + c_{i u_1} +\sum_{j=2}^{L} c_{u_{j-1} u_j} + c_{u_L 0} + \sum_{j=1}^{L} \Big( \sum_{\omega\in\Omega} \alpha^\omega \cdot \pi(a_{\theta'}(u_j; \mathbf{t}^\omega) - \ell_{u_j}) - \gamma_{u_j} \Big) + \widehat{T}_{\textsc{RSCP}} (u_L, Q - q_{\theta'}) \nonumber\\
	\labelrel={prop-ineq-rcsp-5}\,& \overline{C}_{\theta'} + \widehat{T}_{\textsc{RSCP}} (u_L, Q - q_{\theta'}) \,\labelrel\leq{prop-ineq-rcsp-6}\, \overline{C}_{\theta'}. \nonumber 
\end{align*}
Inequality~\eqref{prop-ineq-rcsp-1} follows from the fact that extending path $\theta$ to customer $u_1$ cannot lead to a smaller bound than the bound associated with the optimal path extension from the minimization operator in Equation~\eqref{eq:rcsp-bound-def}.
In Inequality~\eqref{prop-ineq-rcsp-2}, the same argument holds when extending path $\theta\oplus u_1$ to customers $u_2, \dots, u_j$.
In Equation~\eqref{prop-ineq-rcsp-3} we rearrange the terms, and Inequality~\eqref{prop-ineq-rcsp-4} is due to the fact that, given a path $\theta$ ending at customer~$j$:
\begin{equation}
	\delta_\theta \leq \tau_\theta \leq a_\theta (j; \mathbf{t}^\omega), \quad \forall \omega\in\Omega
\end{equation}
Finally, Equality~\eqref{prop-ineq-rcsp-5} is due to the resource extension function for the reduced cost given by Equation~\eqref{eq:label-ext-cost}, and  Inequality~\eqref{prop-ineq-rcsp-6} holds since going from $u_L$ back to the depot incurs no additional cost and can not improve the completion bound.
\hfill$\square$

\vspace{0.2cm}
\textit{Proof of Proposition~\ref{prop:completion-bound-knapsack}.}\hspace*{0.2cm} From the resource extension function given by Equation~\eqref{eq:label-ext-cost}, we have:
\begin{align*}
	\overline{C}_{\theta'} & \labelrel={prop-ineq-knapsack-0} \overline{C}_{\theta} - c_{i0} + c_{i u_1} + \sum_{j=2}^{L} c_{u_{j-1} u_j} + c_{u_L 0} + \sum_{j=1}^{L} \bigg( \sum_{\omega\in\Omega} \alpha^\omega \cdot \pi(a_{\theta'}(u_j; \mathbf{t}^\omega) - \ell_{u_j}) - \gamma_{u_j} \bigg) \\
	&\labelrel\geq{prop-ineq-knapsack-1} \overline{C}_{\theta} + \sum_{j=1}^{L} \bigg( \sum_{\omega\in\Omega} \alpha^\omega \cdot \pi(a_{\theta'}(u_j; \mathbf{t}^\omega) - \ell_{u_j}) - \gamma_{u_j} \bigg) \\
	&\labelrel\geq{prop-ineq-knapsack-2} \overline{C}_{\theta} + \sum_{j=1}^{L} \Big( \pi(\tau_\theta + \min_{\omega\in\Omega} t_{i u_j}^\omega - \ell_{u_j}) - \gamma_{u_j} \Big) \\
	&\labelrel={prop-ineq-knapsack-3} \overline{C}_{\theta} + \sum_{l\in\mathcal{V}\setminus\{0\}} - v_{i l} (\theta)\, z^*_{l} \,\labelrel\geq{prop-ineq-knapsack-4}\, \overline{C}_{\theta} + \widehat{T}_{ks}(i, Q-q_\theta) \,\geq\, 0 
\end{align*}
where Inequality~\eqref{prop-ineq-knapsack-1} is due to the triangle inequality, which implies that the cost of a route cannot decrease if we add customers to it. Inequality~\eqref{prop-ineq-knapsack-2} is a consequence of Equation~\eqref{knapsack-value-bound} and the fact that adding more customers to a route between $i$ and $u_j$ can only increase the arrival time at customer $u_j$. 
Equality~\eqref{prop-ineq-knapsack-3} holds by our definition of the knapsack values.
Inequality~\eqref{prop-ineq-knapsack-4} is due to the optimality of the knapsack solution and the definition of the completion bound. 
\hfill$\square$

\section{Detailed results for instances with 25 customers}\label{sec:results-25}

The \gls{BPC} method could find optimal solutions for all instances with 25 customers within the predetermined time limit. In the following, we analyze the test cost results under the different generative models of travel times.

\vspace*{0.3cm}
\noindent\textbf{Linear generative model.} Table~\ref{res-linear-25} reports the test costs of different data-driven approaches for instances with 25 customers and a linear generative model. Among the practical models, we highlight with boldface numbers the models that achieve the lowest cost on each instance.
For PTO-F and ``Full'' models, we highlight the results in boldface whenever one of the practical models achieves equal or lower test cost.
\gls{CSAA} achieves the lowest test costs among the practical models for most instances. For C-type instances, most models achieve the full information lower bound.

\vspace*{0.3cm}
\noindent\textbf{Exponential generative model.} Table~\ref{res-exponential-25} reports the test costs for instances with 25 customers and an exponential generative model. We observe higher deviations from the lower bound in this setting than in the previous results under the linear generative model. On average, \gls{CSAA} is superior to the other models. We also note that P-NN is superior on a number of instances. However, the lower bound is achieved only in a single instance, illustrating the greater difficulty of solving the contextual stochastic \gls{VRPTW} under a nonlinear generative model.

\vspace*{0.3cm}
\noindent\textbf{Sigmoidal generative model.} Table~\ref{res-sigmoidal-25} reports test costs for instances with 25 customers and a sigmoidal generative model. 
As before, on average, \gls{CSAA} is superior to the other models. For some instances, \gls{CSAA} is outperformed by \gls{SAA}. In this setting, \gls{SAA} is superior to \acrshort{PTO} and P-NN, in contrast to the results with linear and exponential generative models. 

\vspace*{0.3cm}
\noindent\textbf{Comparison of generative models.} Table~\ref{res-ttmodel-25} compares the average test costs of all considered approaches under the different generative models of travel times. Considering the practical models, \gls{CSAA} achieves the lowest test costs on average. 
PTO-F is superior to \gls{CSAA} for the exponential and sigmoidal instances, indicating that the \acrshort{PTO} framework could be the best approach in those settings, provided that one can train a model that can predict the conditional expected travel times perfectly. Further investigation is needed to assess how well the \acrshort{PTO} framework can perform in practice under different predictive models.
However, in all settings, the average test cost performance of \gls{CSAA} is close to that of PTO-F, meaning that \gls{CSAA} would be on par with PTO-F even in the unlikely case in which one could predict the true conditional expected travel times.

\begin{table}[H]
\footnotesize
\centering
\caption{Test cost results for instances with linear generative model and 25 customers}
\label{res-linear-25}
\begin{tabularx}{\textwidth}{X|LLLLLLLL|LL}
\toprule
 Instance & D-avg &  SAA & PTO-OLS & PTO-$k$NN & SAA-$k$NN &  CSAA & RSAA & P-NN & PTO-F & Full (Abs.) \\
\midrule
  R101 &  2.88 &  1.48 &  1.03 &  1.63 &  1.06 &  0.09 & \textbf{0.05} & 1.68 &  1.03 &  649.6 \\
  R102 &  2.13 &  2.46 & \textbf{0.02} &  1.27 &  1.04 & \textbf{0.02} & \textbf{0.02} & 0.56 & 0.02 &  576.5 \\
  R103 &  8.62 &  0.17 &  2.69 &  2.77 & \textbf{0.00} &  0.06 &  0.09 & 0.09 & 0.00 &  465.4 \\
  R104 & 13.98 &  2.94 &  1.29 &  5.76 &  0.34 & \textbf{0.00} & \textbf{0.00} & 0.02 & 0.8 &  435.6 \\
  R105 &  10.5 & 3.4 &  0.28 &  8.13 &  2.51 & \textbf{0.06} & \textbf{0.06} & 0.07 &  0.28 &  544.9 \\
  R106 &  75.0 & \textbf{0.00} &  1.89 & 42.85 &  8.32 & \textbf{0.00} &  0.84 & 2.67 &  5.85 &  502.4 \\
  R107 &  0.16 &  0.16 &  0.25 &  0.25 &  0.91 & \textbf{0.00} & \textbf{0.00} &  0.3 &  0.25 &  437.6 \\
  R108 &  5.34 &  6.12 & 0.9 &  3.37 & 0.9 &  0.17 & \textbf{0.05} & 0.34 &  0.85 &  411.9 \\
  R109 &  1.34 &  1.98 & \textbf{0.02} &  0.75 &  0.77 & \textbf{0.02} & \textbf{0.02} &  \textbf{0.02} & 0.02 &  455.0 \\
  R110 & 22.39 &  4.51 &  0.63 &  4.03 & 2.2 & \textbf{0.04} & \textbf{0.04} & 0.17 &  0.41 &  458.7 \\
  R111 &  108.97 & 12.23 &  0.89 & 31.46 & 27.09 & \textbf{0.13} & \textbf{0.13} & 1.02 &  1.44 &  450.4 \\
  R112 & 29.73 &  5.17 &  1.73 & 29.73 &  5.87 &  0.12 & \textbf{0.02} &  0.1 & 0.9 &  410.3 \\
  C101 & \textbf{0.00} & \textbf{0.00} & \textbf{0.00} & \textbf{0.00} & 3.1 & \textbf{0.00} & \textbf{0.00} &  \textbf{0.00} & 0.00 &  196.5 \\
  C102 & \textbf{0.00} & \textbf{0.00} & \textbf{0.00} & \textbf{0.00} &  0.77 & \textbf{0.00} & \textbf{0.00} & 0.92 & 0.00 &  195.8 \\
  C103 & - & - & - & - & - & - & - &  - & - &  - \\
  C104 & \textbf{0.00} & \textbf{0.00} & \textbf{0.00} & \textbf{0.00} & \textbf{0.00} & \textbf{0.00} & \textbf{0.00} & 0.21 & 0.00 &  187.5 \\
  C105 & \textbf{0.00} & \textbf{0.00} & \textbf{0.00} & \textbf{0.00} & \textbf{0.00} & \textbf{0.00} & \textbf{0.00} &  \textbf{0.00} & 0.00 &  191.8 \\
  C106 & \textbf{0.00} & \textbf{0.00} & \textbf{0.00} & \textbf{0.00} & \textbf{0.00} & \textbf{0.00} & \textbf{0.00} &  \textbf{0.00} & 0.00 &  194.3 \\
  C107 & \textbf{0.00} & \textbf{0.00} & \textbf{0.00} & \textbf{0.00} & \textbf{0.00} & \textbf{0.00} & \textbf{0.00} &  \textbf{0.00} & 0.00 &  191.8 \\
  C108 & \textbf{0.00} & \textbf{0.00} & \textbf{0.00} & \textbf{0.00} & \textbf{0.00} & \textbf{0.00} & \textbf{0.00} &  \textbf{0.00} & 0.00 &  191.8 \\
  C109 & \textbf{0.00} & \textbf{0.00} & \textbf{0.00} & \textbf{0.00} & \textbf{0.00} & \textbf{0.00} & \textbf{0.00} &  \textbf{0.00} & 0.00 &  191.8 \\
 RC101 & 30.89 &  1.73 &  5.96 &  45.5 & 12.68 & \textbf{0.02} &  0.04 & 4.03 & 5.7 &  533.2 \\
 RC102 &  56.1 &  17.4 &  1.43 & 42.96 &  2.68 &  0.08 & \textbf{0.03} & 0.54 &  1.94 &  392.5 \\
 RC103 & 15.55 & 15.17 &  2.67 & 12.67 &  0.12 & \textbf{0.03} & \textbf{0.03} & 1.74 &  2.82 &  344.0 \\
 RC104 &  2.88 &  0.94 &  2.88 &  5.73 &  1.72 & \textbf{0.38} &  0.44 & 2.66 &  0.63 &  319.3 \\
 RC105 &  2.67 &  1.24 &  1.38 &  1.15 & \textbf{0.48} &  0.55 &  0.55 & 1.34 & 0.6 &  419.1 \\
 RC106 & 16.89 &  7.54 &  5.41 & 10.76 &  6.31 &  1.86 & \textbf{1.65} & 6.79 &  2.48 &  375.4 \\
 RC107 & 10.93 & 1.3 &  3.97 &  9.33 &  0.91 &  0.29 & \textbf{0.16} & 3.09 &  2.63 &  307.5 \\
 RC108 &  2.68 &  2.68 & \textbf{0.07} &  3.74 &  0.26 &  0.13 &  0.13 & 0.33 & 0.07 &  302.3 \\
 \midrule 
 Average & 14.99 &  3.16 &  1.26 &  9.42 &  2.86 & \textbf{0.14} &  0.16 & 1.02 &  1.03 & 369.03 \\
\bottomrule
\end{tabularx}
\end{table}

\begin{table}[H]
	\footnotesize
	\centering
	\caption{Test cost results for instances with exponential generative model and 25 customers}
	\label{res-exponential-25}
		\begin{tabularx}{\textwidth}{X|LLLLLLLL|LL}
			\toprule
			Instance &  D-avg & SAA & PTO-OLS & PTO-$k$NN & SAA-$k$NN & CSAA & RSAA & P-NN &  PTO-F & Full (Abs.) \\
			\midrule
			R101 & 8.3 &  4.08 & \textbf{0.33} &  3.29 &  2.49 & 0.7 & 0.52 &  \textbf{0.33} &  0.24 & 1259.2 \\
			R102 &  8.21 & 11.71 &  2.74 &  5.81 &  5.13 & \textbf{1.22} & 1.33 & 1.99 &  0.92 &  779.6 \\
			R103 & 34.77 &  8.01 &  4.05 & 15.32 &  6.94 &  1.32 &  \textbf{1.03} & 2.93 & 1.7 &  522.9 \\
			R104 & 14.99 &  3.11 &  7.26 & 13.58 &  8.48 &  3.74 &  \textbf{2.76} &  4.5 &  3.01 &  510.9 \\
			R105 & 51.75 & 25.06 &  4.98 & 10.67 &  9.76 & \textbf{4.11} & 4.15 &  10.64 &  1.99 &  804.5 \\
			R106 &  7.61 &  8.06 & 1.3 &  2.56 &  2.27 &  1.91 & 1.19 &  \textbf{0.54} &  1.08 &  538.6 \\
			R107 &  356.78 & 46.89 &  3.43 & 20.96 &  6.54 &  0.86 &  \textbf{0.34} &  0.6 &  3.51 &  498.2 \\
			R108 & 34.38 & 10.64 &  3.67 & 12.74 & 4.2 & \textbf{0.49} & 0.79 &  1.2 &  3.67 &  466.3 \\
			R109 &  103.98 & 21.15 &  5.13 & 14.67 &  2.87 & \textbf{2.73} & 3.75 & 2.78 &  2.46 &  557.5 \\
			R110 & 25.27 & 4.3 &  4.32 &  5.36 &  1.96 & \textbf{1.57} & 1.98 & 1.63 &  2.14 &  509.4 \\
			R111 & 31.54 &  23.4 &  1.43 & 30.21 & 12.38 &  1.26 &  \textbf{0.99} & 1.35 &  1.65 &  586.2 \\
			R112 &  103.67 & 14.45 &  5.89 & 53.55 & 13.41 &  6.93 &  6.0 &  \textbf{1.93} &  3.83 &  460.1 \\
			C101 &  176.85 &  8.42 & 47.02 &  178.59 & 39.88 &  4.33 &  5.6 &  \textbf{1.66} &  9.31 &  258.8 \\
			C102 & 33.29 & 6.4 &  5.75 & 20.71 &  4.32 &  1.21 &  \textbf{0.86} & 2.51 &  8.26 &  231.3 \\
			C103 & 20.52 & \textbf{0.35} & 12.53 & 11.43 &  4.49 &  1.95 &  0.9 & 1.75 &  6.54 &  200.3 \\
			C104 & - & - & - & - & - & - &  - &  - & - &  - \\
			C105 &  104.16 & 13.72 & 26.16 &  104.16 & 13.16 & 13.34 & 17.5 &  \textbf{8.32} & 35.03 &  233.2 \\
			C106 & 19.29 &  2.05 & 9.0 & 22.45 &  5.39 & \textbf{0.84} & 2.05 & 2.16 &  3.49 &  263.3 \\
			C107 & \textbf{0.78} & \textbf{0.78} & \textbf{0.78} & \textbf{0.78} & \textbf{0.78} & \textbf{0.78} &  \textbf{0.78} &  \textbf{0.78} & 0.78 &  192.8 \\
			C108 & \textbf{0.10} & \textbf{0.10} & \textbf{0.10} & \textbf{0.10} & \textbf{0.10} & \textbf{0.10} &  \textbf{0.10} &  \textbf{0.10} & 0.10 &  192.7 \\
			C109 & \textbf{0.00} & \textbf{0.00} & \textbf{0.00} & \textbf{0.00} & \textbf{0.00} & \textbf{0.00} &  \textbf{0.00} &  \textbf{0.00} & 0.00 &  191.8 \\
			RC101 & 91.45 &  78.9 & 20.86 & 25.11 & 26.11 & 22.34 & \textbf{20.85} &  20.93 &  2.67 &  674.5 \\
			RC102 &  126.85 & 39.67 &  16.3 & 119.7 &  4.66 &  4.32 & 4.45 &  \textbf{2.88} &  2.02 &  573.5 \\
			RC103 & 42.63 &  6.33 &  9.85 & 14.79 & \textbf{4.94} &  5.81 & 5.81 &  11.14 &  5.37 &  497.5 \\
			RC104 & 53.44 & 19.32 & \textbf{4.91} &  7.56 & 14.94 &  5.92 & 5.87 & 5.75 &  6.95 &  415.6 \\
			RC105 & 15.94 &  5.85 &  4.62 & 15.31 & 4.5 & \textbf{2.44} & 2.71 & 4.84 &  5.76 &  553.8 \\
			RC106 & 350.2 & 21.77 & 10.36 & 27.22 &  6.61 & \textbf{2.07} & 5.06 & 7.42 &  5.55 &  537.4 \\
			RC107 & 11.18 & \textbf{5.46} & 26.13 & 11.67 &  5.87 & 10.59 & 22.0 &  21.92 & 10.63 &  470.4 \\
			RC108 &  106.85 & 20.28 & 12.87 &  53.0 & 16.62 & \textbf{8.54} &  9.9 & 9.29 &  9.27 &  397.0 \\
			\midrule 
			Average &  69.1 & 14.65 &  8.99 & 28.62 &  8.17 & \textbf{3.98} & 4.62 & 4.71 &  4.93 & 477.76 \\
			\bottomrule
	\end{tabularx}
\end{table}
\begin{table}[H]
\footnotesize
\centering
\caption{Test cost results for instances with sigmoidal generative model and 25 customers}
\label{res-sigmoidal-25}
\begin{tabularx}{\textwidth}{X|LLLLLLLL|LL}
\toprule
  Instance &  D-avg & SAA & PTO-OLS & PTO-$k$NN & SAA-$k$NN &  CSAA & RSAA & P-NN &  PTO-F & Full (Abs.) \\
\midrule
R101 &  14.06 & 11.3 &  13.04 &  15.14 & 10.2 &  \textbf{6.31} & 7.34 &  11.27 & 8.57 & 1100.6 \\
  R102 & 123.88 &  30.19 &  25.34 &  79.38 &  39.03 &  21.92 & \textbf{14.00} &  16.09 & 9.44 &  926.2 \\
  R103 & 111.72 &  19.45 &  26.71 &  65.46 &  12.78 &  10.55 &  \textbf{9.26} &  19.47 &  13.97 &  529.0 \\
  R104 &  102.4 &  16.32 &  38.33 &  61.99 &  27.79 &  17.44 & \textbf{14.36} &  27.87 &  21.28 &  500.1 \\
  R105 & 163.67 & 27.7 &  35.37 &  37.17 &  21.83 &  26.02 & \textbf{18.68} &  25.07 &  12.17 &  702.8 \\
  R106 &  55.21 &  12.21 &  56.77 & 49.9 &  12.03 & \textbf{10.34} &  11.69 &  25.38 &  14.79 &  572.5 \\
  R107 & 106.36 & \textbf{13.47} & 91.1 &  76.81 &  16.12 &  15.57 &  13.74 &  36.45 &  28.34 &  514.4 \\
  R108 & 118.65 &  \textbf{8.65} &  40.72 &  86.72 &  13.81 &  17.15 &  11.27 &  24.16 &  15.58 &  473.0 \\
  R109 &  81.88 &  53.94 &  54.22 & 106.62 &  30.46 & \textbf{11.85} &  15.78 &  31.13 &  21.04 &  567.9 \\
  R110 & 180.97 & 12.7 &  20.84 &  42.72 &  11.09 &  \textbf{8.16} &  10.46 &  13.74 &  18.81 &  508.7 \\
  R111 &  24.16 &  11.01 &  20.08 &  20.39 & 14.9 &  \textbf{7.82} & 8.13 &  16.33 &  14.68 &  495.9 \\
  R112 &  88.22 &  12.28 &  46.23 & 63.7 &  17.31 & \textbf{11.27} &  19.93 &  29.36 &  24.87 &  455.1 \\
  C101 & 189.53 & \textbf{12.97} & 137.29 &  67.08 &  14.62 &  21.97 &  18.33 &  39.35 & 127.22 &  272.2 \\
  C102 &  35.91 & \textbf{17.81} &  53.93 &  44.82 &  19.52 & \textbf{17.81} &  19.77 &  32.02 &  69.02 &  239.2 \\
  C103 &  79.54 &  10.61 &  53.04 &  63.37 &  15.79 &  \textbf{2.62} & 7.42 & 8.56 &  29.73 &  210.2 \\
  C104 &  - &  - &  - &  - &  - &  - &  - &  - &  - &  - \\
  C105 & 191.87 & \textbf{14.78} &  181.6 & 186.98 & 136.84 & 132.73 & 134.21 & 145.46 & 154.58 &  243.5 \\
  C106 &  10.33 & 8.43 &  49.47 &  38.37 &  \textbf{7.88} &  16.09 &  23.27 &  33.13 &  91.13 &  272.9 \\
  C107 & 101.25 &  101.2 & \textbf{43.13} & 101.25 & 101.34 &  59.64 &  59.64 & \textbf{43.13} & 101.25 &  216.3 \\
  C108 &  45.79 &  45.79 &  45.79 &  45.79 & \textbf{39.76} &  45.79 &  45.79 &  45.79 &  45.79 &  204.2 \\
  C109 & \textbf{31.37} & \textbf{31.37} & \textbf{31.37} & \textbf{31.37} & \textbf{31.37} & \textbf{31.37} & \textbf{31.37} & \textbf{31.37} & 31.37 &  191.9 \\
 RC101 &  52.94 &  46.07 &  60.23 &  61.72 &  35.21 &  37.71 & 35.0 & \textbf{31.80} &  7.6 &  783.1 \\
 RC102 &  90.07 & \textbf{19.06} & 105.58 &  92.49 &  24.62 &  51.48 & 39.6 &  44.91 &  64.77 &  558.9 \\
 RC103 & 175.45 & \textbf{25.33} &  64.98 &  87.81 &  33.82 &  33.37 &  39.18 &  53.55 &  36.04 &  448.9 \\
 RC104 & 223.16 &  49.96 &  70.28 &  96.86 &  36.56 & \textbf{29.07} & 39.9 &  60.37 &  62.82 &  404.5 \\
 RC105 & 285.53 & \textbf{21.61} &  86.97 & 162.95 & 31.7 &  25.52 &  40.26 &  46.15 &  47.67 &  571.6 \\
 RC106 & 320.05 & 27.7 &  83.45 &  66.89 &  17.76 & \textbf{15.04} &  24.95 &  36.59 &  64.34 &  471.4 \\
 RC107 & 432.47 & \textbf{29.59} & 111.35 & 245.93 &  40.74 &  37.07 & \textbf{29.59} &  61.34 &  47.04 &  485.3 \\
 RC108 & 285.87 & \textbf{29.62} & 129.35 & 149.61 &  54.36 &  32.88 &  55.46 &  84.66 &  68.14 &  410.6 \\
 \midrule 
 Average & 132.94 & \textbf{25.75} &  63.45 &  80.33 &  31.04 &  26.95 &  28.51 &  38.38 &  44.72 &  476.1 \\
\bottomrule
\end{tabularx}
\end{table}

\begin{table}[htbp]
\footnotesize
\centering
\caption{Average test costs and percentage gaps to the full-knowledge model for instances with 25 customers}
\label{res-ttmodel-25}
\begin{tabularx}{\textwidth}{c|m{1.7cm}|LLLLLLLL|LL}
\toprule
& Generative model &  D-avg &  SAA &  PTO-OLS & PTO-$k$NN & SAA-$k$NN & CSAA & RSAA & P-NN &  PTO-F & Full \\
\midrule
\multirow{3}{*}{$\hat{R}$} & Linear &  436.0 &  382.1 &  374.4 &  411.0 &  381.8 & \textbf{369.6} & 369.7 &  373.3 &  373.6 &  369.0 \\
  & Exponential & 809.5 &  559.0 & 515.5 &  596.6 &  515.8 & \textbf{497.3} & 499.8 & 502.0 &  496.5 &  477.8 \\
 &  Sigmoidal &  1102.7 &  593.2 &  751.4 &  844.7 &  607.1 &  \textbf{588.9} & 592.7 & 638.9 &  642.0 &  476.1 \\
\midrule
\multirow{3}{*}{Gap(\%)}  & Linear & 18.16 & 3.55 & 1.46 & 11.38 & 3.47 & \textbf{0.16} & 0.17 & 1.17 & 1.25 & 0.00\\
& Exponential & 69.42 & 16.99 & 7.89 & 24.86 & 7.95 & \textbf{4.08} & 4.61 & 5.06 & 3.91 & 0.00\\
& Sigmoidal & 131.61 & 24.60 & 57.82 & 77.42 & 27.52 & \textbf{23.69} & 24.49 & 34.19 & 34.85 & 0.00\\
\bottomrule
\end{tabularx}
\end{table}

\section{Detailed results for instances with 50 customers}\label{sec:results-50}

With 50 customers, we could not solve all instances to optimality. To conduct a fair comparison between the models, we consider in the following analyses only instances for which all models could find optimal solutions or integer feasible solutions with an optimality gap of at most 0.01.

\vspace*{0.3cm}
\noindent\textbf{Linear generative model.} Table~\ref{res-linear-50} shows test cost results for 22 instances that could be solved to an optimality gap of at most 1\%. Among the practical models, \gls{CSAA} achieves the lowest test costs in all instances. The full-information lower bound is achieved in 5 out of the 22 instances. Notably, the test costs of \gls{CSAA} are smaller than or equal to those of PTO-F in all instances.

\vspace*{0.3cm}
\noindent\textbf{Exponential generative model.} Table~\ref{res-exponential-50} reports the test costs of different data-driven approaches for 24 instances that could be solved to an optimality gap of at most 1\%. For most instances, \gls{CSAA} has the lowest test costs among the practical models. Moreover, \gls{CSAA} is also superior to PTO-F, on average.

\vspace*{0.3cm}
\noindent\textbf{Sigmoidal generative model.} Table~\ref{res-sigmoidal-50} reports the test cost of different data-driven approaches for 21 that could be solved to an optimality gap of at most 1\%. As before, on average, \gls{CSAA} is superior to the other practical models. However, we observe a larger gap to the full-information lower bound.

\vspace*{0.3cm}
\noindent\textbf{Comparison of generative models.} We compare the average test costs achieved by the different methods across the different generative models in Table~\ref{res-ttmodel-50}. For linear and exponential travel times, \gls{CSAA} has the lowest average test costs. For sigmoidal travel times, we observe significantly larger gaps. In this setting, \gls{CSAA} has a full-information gap of 32.93\%, close to PTO-F, with 32.20\%.

\begin{table}[htbp]
\footnotesize
\centering
\caption{Test cost results for instances with linear generative model and 50 customers}
\label{res-linear-50}
\begin{tabularx}{\textwidth}{X|LLLLLLL|LL}
\toprule
  Instance &  D-avg &  SAA &  PTO-OLS &  PTO-$k$NN &  SAA-$k$NN &  CSAA & RSAA &  PTO-F & Full \\
\midrule
R101 & 25.36 &  5.84 &  1.11 & 10.97 &  3.48 & 0.7 & \textbf{0.66} &  0.97 & 1237.0 \\
  R102 & 18.32 &  3.72 &  0.99 & 16.39 &  1.44 & \textbf{0.16} &  0.21 &  1.01 & 1155.8 \\
  R103 &  6.96 &  2.91 &  3.05 &  4.53 &  1.99 & \textbf{0.27} &  0.42 &  1.29 &  850.2 \\
  R104 & - & - & - & - & - & - & - & - &  - \\
  R105 & 51.25 & 6.4 &  2.44 & 23.05 &  5.86 &  0.67 & \textbf{0.28} &  1.28 &  994.4 \\
  R106 & - & - & - & - & - & - & - & - &  - \\
  R107 &  7.52 &  2.12 &  1.81 & 16.82 &  2.63 & 0.3 & \textbf{0.18} &  2.12 &  772.8 \\
  R108 & - & - & - & - & - & - & - & - &  - \\
  R109 &  7.64 &  3.98 &  4.16 &  7.75 &  0.52 & \textbf{0.23} &  0.24 &  1.81 &  828.7 \\
  R110 &  35.0 &  3.22 &  1.45 & 14.33 &  5.92 &  0.44 & \textbf{0.22} &  0.99 &  758.5 \\
  R111 & 32.81 & 3.6 &  2.26 &  19.8 &  6.85 &  0.12 & \textbf{0.08} &  1.58 &  738.5 \\
  R112 & - & - & - & - & - & - & - & - &  - \\
  C101 &  1.66 & 21.32 &  1.66 &  1.66 &  3.09 & \textbf{0.00} & \textbf{0.00} &  1.66 &  379.0 \\
  C102 & 4.7 & 4.7 & \textbf{0.05} & 4.7 &  1.55 & \textbf{0.05} & \textbf{0.05} & \textbf{0.05} &  367.8 \\
  C103 & - & - & - & - & - & - & - & - &  - \\
  C104 & - & - & - & - & - & - & - & - &  - \\
  C105 & \textbf{0.00} & \textbf{0.00} & \textbf{0.00} & \textbf{0.00} & \textbf{0.00} & \textbf{0.00} & \textbf{0.00} & \textbf{0.00} &  363.2 \\
  C106 &  1.52 &  9.57 &  1.52 &  1.52 &  1.52 & \textbf{0.95} & \textbf{0.95} &  1.52 &  387.5 \\
  C107 & \textbf{0.00} & \textbf{0.00} & \textbf{0.00} & \textbf{0.00} & \textbf{0.00} & \textbf{0.00} & \textbf{0.00} & \textbf{0.00} &  363.2 \\
  C108 & \textbf{0.00} & \textbf{0.00} & \textbf{0.00} & \textbf{0.00} & \textbf{0.00} & \textbf{0.00} & \textbf{0.00} & \textbf{0.00} &  363.2 \\
  C109 & \textbf{0.00} & \textbf{0.00} & \textbf{0.00} & \textbf{0.00} & \textbf{0.00} & \textbf{0.00} & \textbf{0.00} & \textbf{0.00} &  363.2 \\
 RC101 & 54.67 & 6.1 & 5.9 & 16.93 &  6.23 &  1.28 & \textbf{1.09} &  4.91 & 1130.4 \\
 RC102 & 93.37 &  9.48 &  3.92 & 24.37 &  6.37 & \textbf{0.75} &  0.81 & 2.4 &  987.5 \\
 RC103 & 45.81 &  5.42 &  4.35 & 22.79 & 11.45 & \textbf{0.10} &  0.46 &  2.13 &  774.5 \\
 RC104 & - & - & - & - & - & - & - & - &  - \\
 RC105 & 47.79 &  2.21 &  7.64 & 40.57 & 17.12 & 0.6 & \textbf{0.44} &  4.59 &  917.1 \\
 RC106 & 27.19 &  5.22 &  4.55 & 17.09 &  3.17 &  0.91 & \textbf{0.65} &  4.77 &  879.3 \\
 RC107 & 56.66 &  5.72 &  1.82 & 18.36 &  2.18 &  0.26 & \textbf{0.09} &  2.71 &  692.7 \\
 RC108 & - & - & - & - & - & - & - & - &  - \\
 \midrule 
 Average & 24.68 &  4.83 &  2.32 & 12.46 &  3.87 &  0.37 & \textbf{0.33} & 1.7 & 728.79 \\
\bottomrule
\end{tabularx}
\end{table}

\begin{table}[htbp]
\footnotesize
\centering
\caption{Test cost results for instances with exponential generative model and 50 customers}
\label{res-exponential-50}
\begin{tabularx}{\textwidth}{X|LLLLLLL|LL}
\toprule
  Instance &  D-avg &  SAA &  PTO-OLS &  PTO-$k$NN & SAA-$k$NN &  CSAA & RSAA &  PTO-F & Full \\
\midrule
R101 & 10.48 & 10.82 &  3.14 &  9.02 &  4.51 & \textbf{1.75} &  2.29 &  1.63 & 1466.0 \\
R102 & 45.24 & 15.37 & 3.4 &  8.94 &  6.75 &  2.81 & \textbf{2.66} &  1.28 & 1821.7 \\
R103 &  20.0 & 18.88 &  5.54 & 11.45 &  9.37 &  3.93 & \textbf{3.92} &  3.05 & 1679.8 \\
R104 & 31.36 &  8.07 & 21.56 & 18.97 & \textbf{3.59} &  8.09 & 13.81 &  6.43 &  794.4 \\
R105 &  4.76 &  4.33 &  7.89 &  3.59 & \textbf{3.01} &  4.72 &  5.33 &  3.36 & 1252.7 \\
R106 & - & - & - & - & - & - & - & - &  - \\
R107 & 198.4 &  23.5 & 13.08 & 34.34 & 11.52 & \textbf{4.12} & 4.6 &  5.21 &  907.8 \\
R108 & - & - & - & - & - & - & - & - &  - \\
R109 & 24.21 & 11.09 &  6.54 &  7.95 &  5.43 & \textbf{2.44} &  3.23 &  5.33 &  975.9 \\
R110 & - & - & - & - & - & - & - & - &  - \\
R111 &  158.52 & 16.74 &  4.38 & 21.72 &  4.78 & \textbf{3.78} &  3.92 &  6.69 &  887.8 \\
R112 & 14.73 & 7.8 &  5.37 & 24.43 &  6.52 &  3.45 & \textbf{3.23} &  6.15 &  756.5 \\
C101 &  4.79 &  1.85 &  6.97 & 4.9 &  2.48 & \textbf{1.65} &  1.69 &  7.05 &  492.3 \\
C102 & 5.7 & \textbf{0.13} &  2.74 &  2.74 &  0.99 &  0.65 &  1.12 &  2.41 &  473.7 \\
C103 & - & - & - & - & - & - & - & - &  - \\
C104 & - & - & - & - & - & - & - & - &  - \\
C105 & 124.2 &  6.62 & 12.26 & 70.73 &  5.23 & \textbf{3.79} &  4.06 & 12.74 &  438.0 \\
C106 & 66.09 &  7.29 &  7.97 & 26.75 &  6.51 &  9.75 & \textbf{5.75} &  7.49 &  500.5 \\
C107 &  113.86 & 60.36 &  3.89 &  113.86 &  7.64 & \textbf{2.64} & \textbf{2.64} &  3.89 &  393.8 \\
C108 & \textbf{0.14} & \textbf{0.14} &  0.63 & \textbf{0.14} & \textbf{0.14} & \textbf{0.14} &  0.22 &  0.69 &  364.0 \\
C109 & \textbf{0.00} & \textbf{0.00} & \textbf{0.00} & \textbf{0.00} & \textbf{0.00} & \textbf{0.00} & \textbf{0.00} & \textbf{0.00} &  363.2 \\
 RC101 &  120.37 & 44.11 & 5.5 & 18.79 & 64.71 & \textbf{3.37} &  4.64 &  2.06 & 1663.0 \\
 RC102 & 96.18 & 49.45 & 14.25 & 24.92 & 25.28 & \textbf{7.31} &  9.02 &  3.93 & 1605.4 \\
 RC103 & 44.07 & 15.06 &  7.25 & 18.17 &  6.69 & \textbf{2.51} &  4.28 & 5.2 & 1029.9 \\
 RC104 & - & - & - & - & - & - & - & - &  - \\
 RC105 & 39.43 & 17.38 & 19.16 & 14.49 &  8.52 & \textbf{4.09} & 10.17 &  5.56 & 1235.6 \\
 RC106 &  109.79 & 55.04 & 13.84 & 46.25 &  35.6 & \textbf{9.23} & 13.02 &  4.87 & 1181.6 \\
 RC107 & 58.82 & \textbf{6.72} & 98.41 & 32.39 &  7.36 & 12.33 &  20.5 &  9.71 &  928.5 \\
 RC108 & - & - & - & - & - & - & - & - &  - \\
 \midrule 
 Average & 58.69 & 17.31 & 11.99 & 23.39 &  10.3 & \textbf{4.21} &  5.46 &  4.76 & 964.19 \\
\bottomrule
\end{tabularx}
\end{table}

\begin{table}[htbp]
\footnotesize
\centering
\caption{Test cost results for instances with sigmoidal generative model and 50 customers}
\label{res-sigmoidal-50}
\begin{tabularx}{\textwidth}{X|LLLLLLL|LL}
\toprule
 Instance & D-avg & SAA & PTO-OLS & PTO-$k$NN & SAA-$k$NN & CSAA & RSAA & PTO-F & Full \\
\midrule
 R101 & 87.09 & 31.48 & 43.26 & 54.85 & 27.75 & \textbf{21.85} & 29.31 & 8.52 & 2179.0 \\
 R102 & 131.73 & 53.93 & 53.77 & 69.1 & 53.27 & \textbf{29.65} & 30.23 & 18.79 & 1732.0 \\
 R103 & 241.33 & 46.09 & 76.05 & 90.64 & \textbf{26.26} & 26.61 & 35.35 & 17.02 & 1627.7 \\
 R104 & - & - & - & - & - & - & - & - & - \\
 R105 & 303.68 & 23.6 & 32.32 & 104.4 & \textbf{16.23} & \textbf{16.23} & 17.99 & 25.09 & 1278.4 \\
 R106 & 252.94 & 43.82 & 106.23 & 107.35 & 42.77 & 28.46 & \textbf{22.93} & 50.4 & 1021.1 \\
 R107 & 233.67 & 32.1 & 92.73 & 81.15 & 31.91 & \textbf{23.07} & 25.51 & 37.18 & 940.9 \\
 R108 & - & - & - & - & - & - & - & - & - \\
 R109 & 261.41 & \textbf{16.81} & 72.51 & 156.4 & 26.91 & 19.46 & 22.79 & 41.33 & 978.7 \\
 R110 & 167.15 & \textbf{18.02} & 80.05 & 107.21 & 35.25 & 26.35 & 23.18 & 44.34 & 862.7 \\
 R111 & 164.21 & 26.28 & 80.91 & 115.21 & 31.36 & \textbf{23.87} & 30.57 & 44.54 & 873.8 \\
 R112 & 170.69 & 28.15 & 83.54 & 144.91 & 37.41 & 29.05 & \textbf{15.94} & 33.91 & 795.3 \\
 C101 & 186.14 & \textbf{16.02} & 59.32 & 62.29 & 25.45 & 49.13 & 29.6 & 63.04 & 532.5 \\
 C102 & 128.54 & \textbf{16.09} & 108.06 & 124.41 & 24.31 & 21.4 & 24.45 & 48.58 & 506.0 \\
 C103 & - & - & - & - & - & - & - & - & - \\
 C104 & - & - & - & - & - & - & - & - & - \\
 C105 & - & - & - & - & - & - & - & - & - \\
 C106 & - & - & - & - & - & - & - & - & - \\
 C107 & 52.86 & 52.86 & 52.68 & 52.86 & 52.46 & \textbf{52.26} & 52.68 & 52.86 & 402.4 \\
 C108 & - & - & - & - & - & - & - & - & - \\
 C109 & 12.25 & 12.25 & 12.25 & \textbf{12.00} & 12.8 & \textbf{12.00} & 12.25 & 12.74 & 364.1 \\
 RC101 & 228.78 & \textbf{33.32} & 89.05 & 80.05 & 36.36 & 46.11 & 53.35 & 9.24 & 2619.1 \\
 RC102 & 292.43 & 97.24 & 88.67 & 168.9 & 79.34 & \textbf{57.24} & 68.13 & 41.78 & 1337.6 \\
 RC103 & 509.92 & 80.08 & 143.89 & 143.32 & 77.7 & \textbf{32.20} & 37.6 & 47.19 & 1021.2 \\
 RC104 & - & - & - & - & - & - & - & - & - \\
 RC105 & 262.02 & \textbf{36.06} & 84.24 & 294.37 & 76.76 & 50.57 & 51.29 & 27.66 & 1422.8 \\
 RC106 & 203.06 & 44.88 & 90.33 & 90.65 & \textbf{35.20} & 41.64 & 39.01 & 72.9 & 1100.6 \\
 RC107 & 477.15 & \textbf{21.88} & 149.12 & 237.37 & 36.3 & 34.21 & 26.74 & 54.2 & 982.0 \\
 RC108 & - & - & - & - & - & - & - & - & - \\
 \midrule 
 Average & 218.35 & 36.55 & 79.95 & 114.87 & 39.29 & \textbf{32.07} & 32.44 & 37.57 & 1128.9 \\
\bottomrule
\end{tabularx}
\end{table}

\begin{table}[htbp]
\footnotesize
\centering
\caption{Average test costs and percentage gaps to the full-knowledge model for instances with 50 customers}
\label{res-ttmodel-50}
\begin{tabularx}{\textwidth}{c|l|LLLLLLL|LL}
\toprule
& Gen. model & D-avg & SAA & PTO-OLS & PTO-$k$NN & SAA-$k$NN & CSAA & RSAA & PTO-F & Full \\
\midrule
\multirow{3}{*}{$\hat{R}$} & Linear & 953.4 & 764.5 & 749.1 & 839.7 & 762.1 & 732.1 & \textbf{731.7} & 743.5 & 728.8 \\
 & Exponential & 1549.8 & 1160.9 & 1082.4 & 1159.9 & 1096.5 & \textbf{1007.3} & 1021.7 & 1006.1 & 964.2 \\
 & Sigmoidal & 3690.0 & 1570.5 & 2029.8 & 2440.4 & 1588.8 & \textbf{1500.7} & 1524.4 & 1492.4 & 1128.9 \\
\midrule
\multirow{3}{*}{Gap(\%)} & Linear & 30.83 & 4.90 & 2.78 & 15.22 & 4.57 & 0.46 & \textbf{0.40} & 2.02 & 0.00 \\
& Exponential & 60.73 & 20.40 & 12.26 & 20.30 & 13.72 & \textbf{4.47} & 5.96 & 4.34 & 0.00 \\ 
& Sigmoidal & 226.87 & 39.12 & 79.80 & 116.18 & 40.73 & \textbf{32.93} & 35.04 & 32.20 & 0.00\\
\bottomrule
\end{tabularx}
\end{table}

\end{document}